\documentclass[conference]{IEEEtran}
\IEEEoverridecommandlockouts
\usepackage{graphicx}
\usepackage{cite}
\usepackage{amsmath,amssymb,amsfonts}
\usepackage{enumerate}
\usepackage{booktabs}  
\usepackage{threeparttable}
\usepackage{multirow}
\usepackage{graphicx}
\usepackage{textcomp}
\usepackage{xcolor}
\usepackage{algorithm}
\usepackage{algpseudocode}
\usepackage{float}
\usepackage{lipsum}



\def\BibTeX{{\rm B\kern-.05em{\sc i\kern-.025em b}\kern-.08em
    T\kern-.1667em\lower.7ex\hbox{E}\kern-.125emX}}

\begin{document}

\newcommand{\blue}[1]{\textcolor{blue}{#1}}
\newcommand{\hw}[1]{\textcolor{blue}{\textit{#1}}}   %Haiyu Comments

\title{Financial ticket intelligent recognition system based on deep learning}

\author{
Fukang Tian$^{1}$, Haiyu Wu$^{1}$, and Bo Xu$^{12}$
\thanks{$^{1}$Xi'an Network Computing Data Technology Co., Ltd.}
\thanks{$^{2}$Corresponding author}
}

\maketitle
% ******************************************************************************** Begin
\begin{abstract}

Facing the rapid growth in the issuance of financial tickets (or bills, invoices etc.), traditional manual invoice reimbursement and financial accounting system are imposing an increasing burden on financial accountants and consuming excessive manpower. To solve this problem, we proposes an iterative self-learning Framework of Financial Ticket intelligent Recognition System (FFTRS), which can support  the fast iterative updating and extensibility of the algorithm model, which are the fundamental requirements for a practical financial accounting system. In addition, we designed a simple yet efficient Financial Ticket Faster Detection network (FTFDNet) and an intelligent data warehouse of financial ticket are designed to strengthen its efficiency and performance. At present, the system can recognize 194 kinds of financial tickets and has an automatic iterative optimization mechanism, which means, with the increase of application time, the types of tickets supported by the system will continue to increase, and the accuracy of recognition will continue to improve. Experimental results show that the average recognition accuracy of the system is 97.07\%, and the average running time for a single ticket is 175.67ms. The practical value of the system has been tested in a commercial application, which makes a beneficial attempt for the deep learning technology in financial accounting work.

% In this paper, we integrate the current deep learning algorithm and propose a Financial Ticket Detection network(FTFDNet) according to the characteristics of the types of tickets. Furthermore, combined with the actual requirements of finance and taxation business, we build a digital ticket-warehouse module, and distinguish two branches of forward call and feedback optimization to build the financial ticket intelligent recognition system. At present, the system could support the recognition task for 482 types of tickets, and have a self-iterative optimization mechanism, which promises the increased types and improved accuracy for recognition with the deepening business. Through experimental observation, the average recognition accuracy for different types of tickets of our system is 97.07\%, and the average single-ticket recognition time is 150.57ms, which meets the requirements of real-time processing and low fault tolerance of finance and taxation business. The practical value has been tested in online business, which makes a beneficial attempt for the in-depth application of deep learning technology in the field of finance and taxation.

\end{abstract}


\section{Introduction}
In recent years, with the rapid development of computer hardware, computer vision and other technologies, deep learning is being adopted by an ever-widening group of fields\cite{miikkulainen2019artificial,feng2019computer,charniak2019introduction,solis2019domain,o2019deep}. Finance-and-tax is an important field that implements deep learning applications\cite{jha2019automation,srivastava2019optical}. Traditionally, accounting is usually performed manually as follows. First, the different types of financial tickets, such as value-added tax (VAT) invoices (common invoices, electronic invoices, and special invoices), bank tickets, toll tickets (highway passenger tickets, vehicle occupation fees, highway tolls,) are manually sorted. Second, the basic information of these financial tickets is manually input into the financial software to produce accounting vouchers for the corresponding category. Then, each financial ticket is sequentially attached to the accounting voucher for the corresponding category. Finally, the accountant must repeatedly check whether the ordering of the tickets is correct and whether there are any missing tickets. However, this approach is obviously slowed by the lack of automation. Due to the large number and variety of financial tickets, the process results in massive classification workloads, time consumption, and labor effort on the part of the accounting staff, leading to high labor costs and low work efficiency. The accuracy of input information is also greatly affected. Therefore, in order to make accounting more accurate, more efficient and highly automated, optical character recognition (OCR) technology has been gradually applied to the field of financial ticket recognition\cite{srivastava2019optical,ha2017recognition,shreya2019optical}. The ticket information identification system can not only reduce work tasks and pressure and improve work efficiency but also resolve contradictions caused by rising labor costs and labor shortages. Additionally, it can promote the digitalization, maintenance and intelligent accounting and storage of accounting information, making it more convenient for accountants to review. At present, some researches have put forward some ticket recognition systems\cite{sun2019template,zhang2019research,cesarini2003analysis,palm2017cloudscan, blanchard2019automatic, yi2019dual,klein2004results, kieri2012context}, but there are some common problems: 1) They could only support a few types of tickets. 2) The accuracy is low, which cannot meet the accuracy requirements of the financial application. 3) Their systems do not support complex scenes. Because the experimental data collection methods are not rich enough, the system does not support the real complex scene. 4) Slow iteration of ticket category update. The main reason for these problems is: the lack of understanding of the characteristics of financial ticket image data in the process of system development.

\begin{figure}[htbp]
%\vskip 0.2in
\begin{center}
\centerline{\includegraphics[width=1\columnwidth]{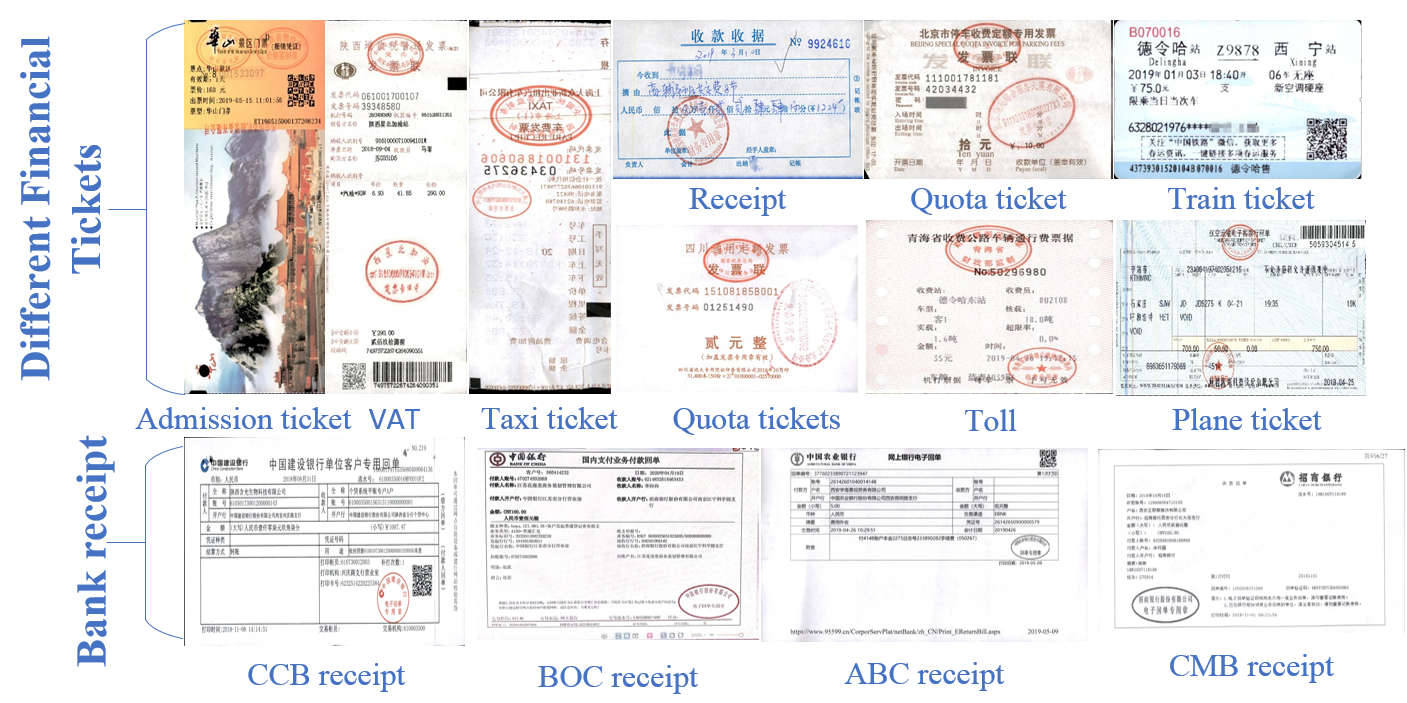}}
\caption{Complex financial tickets}
\label{fig15}
\end{center}
%\vskip -0.2in
\end{figure}

\begin{figure*}[htbp]
%\vskip 0.2in
\begin{center}
\centerline{\includegraphics[width=2\columnwidth, height=8cm]{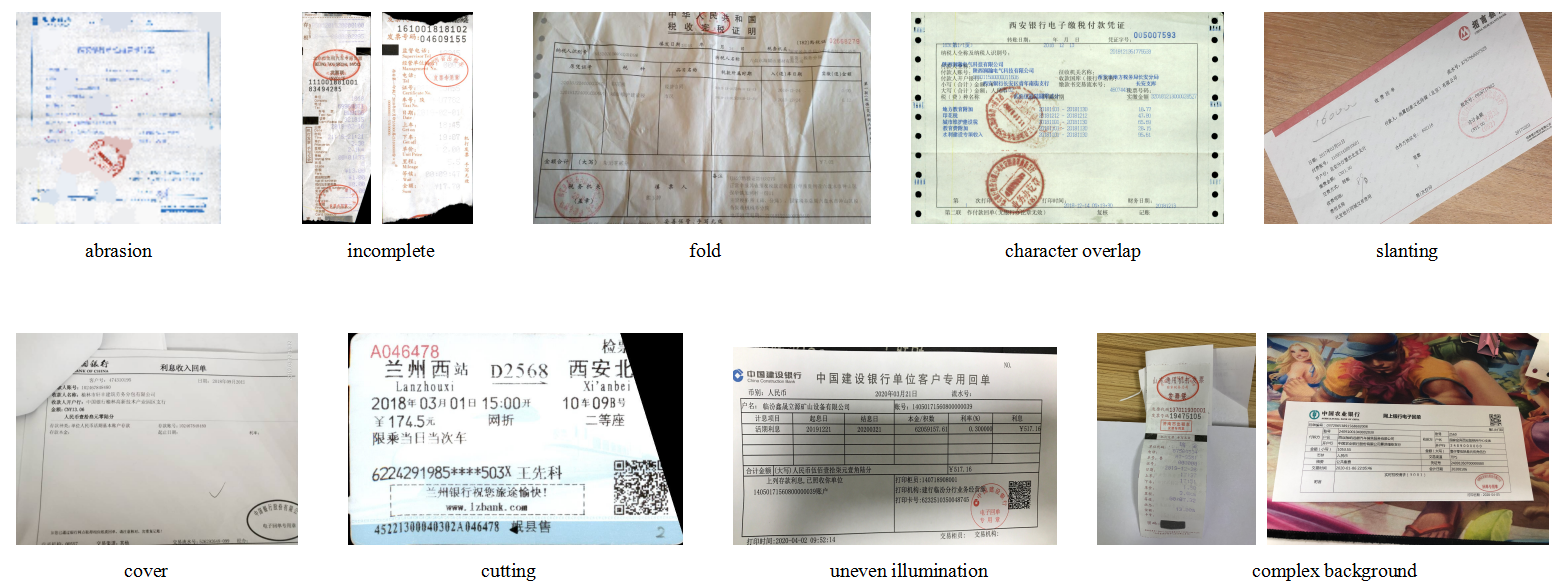}}
\caption{The  quality  of  the  ticket itself and the acquisition equipment, technical differences and accidental factors in the collection process will lead to different ticket  image  quality.}
\label{fig11}
\end{center}
%\vskip -0.2in
\end{figure*}

The image of financial ticket is the image produced by digitizing the financial ticket by scanning and photographing. The characteristics of the financial ticket and the complexity of the image acquisition process make it have two obvious characteristics: 1) As shown in Fig.\ref{fig15}, there are many different types of financial tickets, such as value-added tax (VAT) invoices (common invoices, electronic invoices, and special invoices), bank tickets, toll tickets (highway passenger tickets, vehicle occupation fees, highway tolls,) financial bills, financial receipt, tickets for tourist attractions, admission ticket, plane ticket,train ticket, taxi ticket, toll invoice and so on,  we call all of them tickets, which leads to the complexity of financial tickets. For example, there are 4034 registered banks in China, and each bank uses different ticket styles as different business vouchers. 2) The image quality is uneven. As shown in Fig.\ref{fig11}, the quality of the ticket itself and the acquisition equipment, technical differences and accidental factors in the collection process will lead to different ticket image quality. It mainly includes abrasion, deformity, wrinkle, character overlap, tilt, occlusion, cutting, uneven illumination and complex background.So,how to build an intelligent ticket recognition system with strong adaptability, high precision and high efficiency is a challenge. The problems include: 1) Efficient and accurate ticket classification. 2) Accurate identification of ticket content. 3) Structure of ticket information. 4) Correct accounting entry of ticket. In addition, it also includes the ticket preprocessing processes, such as ticket cutting, correction, enhancement, etc. Traditional methods, image processing\cite{bailey2007single}\cite{sauvola2000adaptive} and machine learning \cite{platt1998sequential}\cite{ratsch2001soft}, gives a great contribution on solving these problems. They improved the ticket classification and recognition accuracy up to 96\% and 85.22\% respectively. However, the practicability of these methods is limited by the following three aspects: 1) Financial software has a very low tolerance for errors, and in many cases, subtle differences lead to very different results. Therefore, when the ticket database is large, the output results cannot meet the requirements of financial software for data accuracy. 2) The traditional algorithm lacks effective solutions to the noise of affine and projection transformation caused by wrinkles, breakage, character overlap, and serious deformation of tickets, which are the main reason causes the low accuracy. 3) The existing methods are limited by the total amount and categories of tickets used in the research, and the types of supporting tickets are not wide enough to meet the needs of financial software systematically. In addition, most of the existing methods fail to carry out specific research on different types of tickets, so there is still room for improving effectiveness and efficiency.

Nowadays, the deep learning algorithm, as the most powerful tool for classification and recognition task, has been implemented in different realms. For detection work, so many extraordinary networks are designed, such as Faster-RCNN\cite{ren2015faster}, Mask-RCNN\cite{he2017mask}, YOLOs\cite{redmon2016you,redmon2017yolo9000,redmon2018yolov3,bochkovskiy2020yolov4}, etc. For the word recognition task, \cite{zhan2019esir,xie2019aggregation,wan2020vocabulary,qiao2020seed} solve the problems on both blurred words recognition and arbitrary-shape words recognition. On the basis of the above technologies, we extracted 482 categories of 1.1 million pieces of financial tickets data from the actual operation of the financial SaaS platform as the experimental dataset, designed and built an intelligent recognition system of financial tickets. Moreover, it has been verified. Our main contributions are as follows:
\begin{itemize}
\item Propose a framework for the financial ticket recognition system. The system framework widely supports the existing types of financial tickets, and can automatically iterate and optimize the error-prone and new types of tickets with the development of business. It provides a reference for the in-depth application of deep learning technology in the financial field.
\item Design a digital ticket-warehouse to automatically update the iterative data model. Through the ticket warehouse, the manual audit and identification model training system are linked to push the ticket real-time monitoring and intelligent training iterative updating model, so that the system has stronger practical value.
\item According to the simple and effective principle, Faster-RCNN based FTFDNet is designed to improve the overall efficiency of the system.
\item Using the actual business data, experiments are carried out from three aspects of time-consuming, precision and computing resource consumption, which fully verifies the practicability of the proposed system, and provides reference for subsequent research.
\end{itemize}

\section{RELATED WORK}
\subsection{Ticket recognition technology}As an important application scenario in the OCR field, ticket recognition has been studied in depth. Traditional OCR technology based on image processing, using artificial features, such as Shannon Entropy\cite{shannon1948mathematical}, different filters\cite{luck1994spatial, badcock1990low, maragos1987morphological}, Otsu’s method\cite{otsu1979threshold}, Niblack’s method\cite{rais2004adaptive}, has realized the rapid detection and recognition of ticket text and achieved obvious results in removing salt and pepper noise and Gaussian noise. However, the recognition accuracy is generally low. The OCR technology using statistical machine learning, such as Adaboost\cite{hastie2009multi}, SVM\cite{platt1998sequential}, has a higher anti-noise ability and significantly improves recognition accuracy. However, this kind of algorithm cannot achieve the expected effect when dealing with the complex but common noise on tickets such as fold, damage, character overlap, and deformation. With the extensive application of deep learning technology in OCR field, the above problems are obviously improved. FasterRCNN, YOLOs\cite{redmon2016you,redmon2017yolo9000,redmon2018yolov3,bochkovskiy2020yolov4}, CTPN\cite{DBLP:journals/corr/TianHHH016}, SegLink\cite{shi2017detecting}, PixelLink\cite{deng2018pixellink}, EAST\cite{zhou2017east}, Textboxes\cite{liao2016textboxes}, Graph Convolutional Network\cite{kipf2016semi} etc. have achieved excellent results in text region detection. CRNN\cite{shi2016end}, RARE\cite{shi2016robust}, FOTS\cite{liu2018fots}, Rotation-Sensitive Regression\cite{liao2018rotation}, and STN-OCR\cite{bartz2017stn} provide more choices for character recognition. These methods have a strong ability to deal with different noises and significantly improve the performance of ticket recognition.
\subsection{Ticket recognition system}With the continuous progress of ticket recognition technology, the ticket recognition system has been developed rapidly. Palm\cite{palm2017cloudscan} uses RNN to implement a ticket analysis system named CloudScan, which does not need prior information of ticket, and the average recall rate reaches 89.1\%.  Sun\cite{sun2019template} proposed a ticket intelligent recognition system by using template matching method. The system uses the prior information of bills to determine the area to be recognized, and achieves 95\% recognition accuracy and 14ms single sample processing speed. Yi\cite{yi2019dual} proposed a ticket recognition system based on Gaussian fuzzy and deep learning model for medical invoice recognition task. The accuracy of the system has been improved. The above systems all use deep learning technology, which improves the accuracy, but at the same time, it also exposes two common problems in the current ticket recognition system. One is that the diversity of training data types is not enough, which leads to insufficient support for tickets types. The other is that the recognition system is not closely combined with financial business, and it can not deal with all aspects of tickets from input to entry. At the same time, the system itself has insufficient ability to update and iterate automatically in engineering business.

\section{Framework and modules}
\begin{figure}[htbp]
%\vskip 0.2in
\begin{center}
\centerline{\includegraphics[width=1\columnwidth]{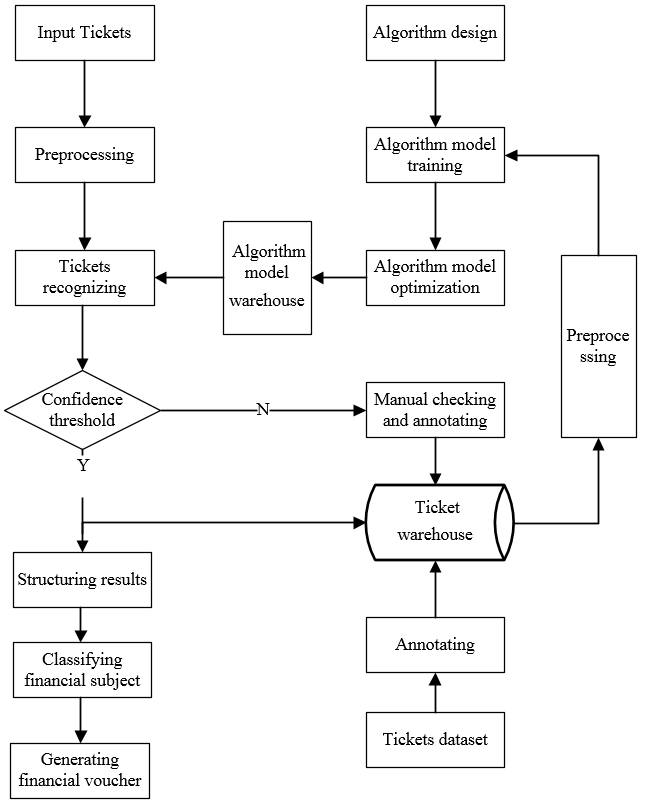}}
\caption{The framework of financial ticket intelligent recognition system}
\label{fig1}
\end{center}
%\vskip -0.2in
\end{figure}

The goal of the intelligent recognition system is to realize the batch extraction of useful information on the financial ticket images that can be used for accounting to carry out financial processing and assist the accountants to complete the financial work. As shown in Fig.\ref{fig1}: the system is composed of image preprocessing module, recognition module, ticket warehouse module and  algorithm model warehouse module. Image preprocessing module and recognition module constitute a forward call function branch, which mainly achieves ticket information extraction and accounting information integration. digital ticket-warehouse provides a database, which is used to store the business data, for forward call branch. Meanwhile, data analysis and transfer, data information perfection, and data intelligent push function in ticket warehouse provide data support for model selection and optimization in the  algorithm model warehouse module. This process constitutes a system feedback optimization function branch.

According to the business requirements, the forward call branch selects the best model from the  algorithm model warehouse to complete the ticket recognition. The recognition result is measured by the confidence threshold. If the recognition result does not reach the threshold value, the auditor intervenes to add verification information. After the ticket passes the forward call branch, on the one hand, it will be used to generate business vouchers according to the identification results. On the other hand, it will be pushed to the ticket warehouse to form data resources. This feedback optimization branch uses the data resources provided by the ticket warehouse, dynamically optimizes each link model according to the actual business problems and data magnitude changes. Sequentially, the optimized model enters the  algorithm model warehouse and waits for the calling of the forward call branch. These two branches are connected by ticket warehouse and  algorithm model warehouse, and the amount of data information is increased by the work of ticket auditors to form a loop to realize cycle optimization. The followings focus on three main modules of the system: ticket warehouse, data preprocessing and ticket recognition.

\subsection{Ticket warehouse}
Ticket warehouse is a database that provides data support for daily business, cycle optimization, and algorithm research of financial ticket intelligent recognition system. In this database, all data samples are divided into four levels according to the existing information shown in Table \ref{table1}.

\begin{table}[!hbt]
\caption{AE stands for accounting entry; KBB stands for keywords bounding box; KC stands for keywords content; TC stands for ticket classification}
\begin{center}
 \begin{tabular}{ccccc}
\hline
  \bfseries information level & \bfseries AE & \bfseries KBB & \bfseries KC & \bfseries TC\\
\hline
\textbf{Level1}&\textbf{\checkmark}&\textbf{\checkmark}&\textbf{\checkmark}&\textbf{\checkmark}\\

\textbf{Level2}&\textbf{}&\textbf{\checkmark}&\textbf{\checkmark}&\textbf{\checkmark}\\

\textbf{Level3}&\textbf{}&\textbf{}&\textbf{\checkmark}&\textbf{\checkmark}\\

\textbf{Level4}&\textbf{}&\textbf{}&\textbf{}&\textbf{\checkmark}\\
\hline
\end{tabular}
\end{center}
\vskip -0.2in
\label{table1}
\end{table}

The ticket warehouse has three functions: A) According to the data information level, automatically store, call, and analyze data. Ticket warehouse can divide the data generated from multi-source in daily business according to the level of information, form a multi-dimensional cross retrieval storage structure, and realize an efficient and accurate data transfer and quantitative analysis. B) It has an open-data information improvement function. Ticket warehouse provides an interactive interface to receive data entered from ticket auditors, data tagging personnel, business customers, developers, and other aspects. It can update data information in real-time and improve the level of data information. C) It has an intelligent data push function. Ticket warehouse can intelligently push error-prone data, unfamiliar data, and scarce data to model candidate training data set depending on the algorithm optimization requirements in the  algorithm model warehouse. The definition of these three data are shown as follows:

\begin{itemize}
\item Error-prone data: When this kind of data passes through the system, the confidence level of classification and recognition is higher than the threshold value, and then it is judged as wrong classification or recognition result manually.
\item Unfamiliar data: When this kind of data passes through the system, the confidence level of classification or recognition is usually lower than the threshold value, and it is judged as a new classification manually.
\item Scarce data: This kind of data is rare, so that the temporary use of various types of data augmentation methods to make up for the problem of insufficient data in training model. In order to make the distribution of training samples closer to the real scene, the ticket warehouse collects such data for a long time.
\end{itemize}

\begin{figure}[htbp]
%\vskip 0.2in
\begin{center}
\centerline{\includegraphics[width=1\columnwidth]{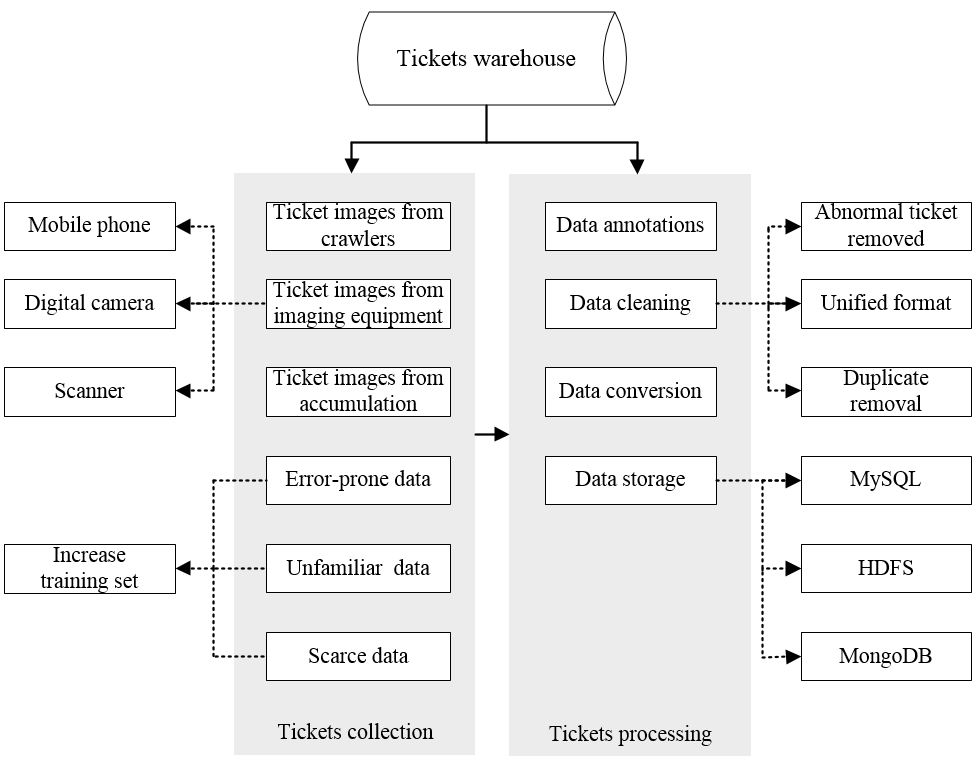}}
\caption{Ticket warehouse module}
\label{fig2}
\end{center}
%\vskip -0.2in
\end{figure}

As shown in Fig.\ref{fig2}, through the above functions, the ticket warehouse completes the real-time storage and sorting of the data stream which is generated by the business of the identification system. In addition, it carries out data statistical analysis on the forward call branches regularly and provides business decision support and algorithm optimization decision suggestions. Moreover, the ticket warehouse continuously provides fresh data for the feedback optimization branch, which ensures the continuous iterative optimization of each algorithm in the system.

\subsection{Tickets preprocessing module}
Ticket image preprocessing in the system refers to region segmentation and direction correction, so that each image entered in the system could maintain a single ticket area, remove the non ticket background and be at the upward angle of the text.

\subsubsection{Region segmentation}
\begin{figure}[htbp]
%\vskip 0.2in
\begin{center}
\centerline{\includegraphics[width=1\columnwidth]{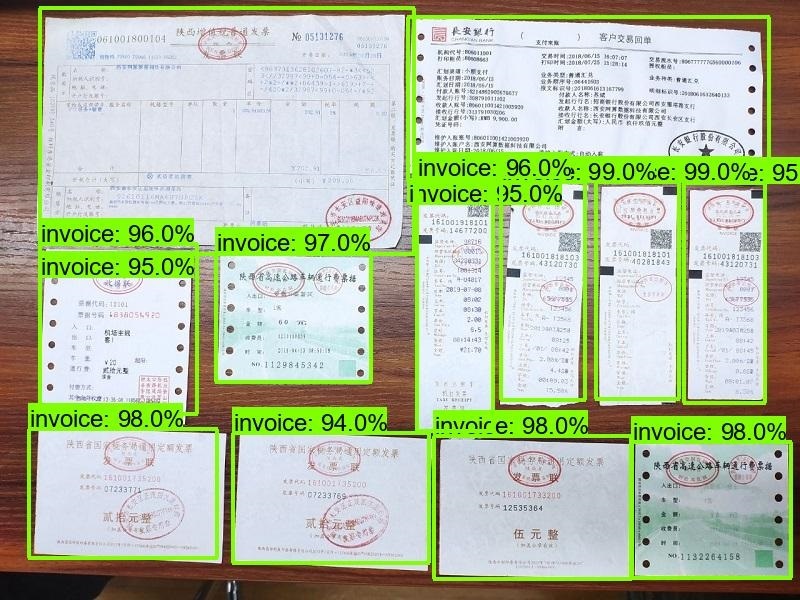}}
\caption{Region segmentation for multiple tickets in one image}
\label{fig3}
\end{center}
%\vskip -0.2in
\end{figure}

There are many sources of financial ticket images, and the collection methods are diverse. Some images contain multiple ticket regions and multiple ticket samples, but some only contain a single ticket area with a large background area remaining. All of these have an impact on the classification, detection, and recognition of ticket images. The essence of ticket region segmentation is to take the ticket area as the target object and detect it from the image.A variety of object detection algorithms can complete this task.The traditional machine learning algorithm for processing artificial design features has the advantages of fast training speed and processing speed, and low requirement for the number of training data sets, but its accuracy is relatively low.The SVM detection algorithm based on HOG feature provided by DLIB and Haar feature based detection algorithm provided by OpenCV are representative.The processing speed of deep learning algorithm is slow,the training difficulty and training data set requirements are relatively high, but its detection accuracy is high.Faster-RCNN and SSD algorithms are representative.Table \ref{table2} shows the comparison among Faster-RCNN, SSD\cite{liu2016ssd}, OpenCV-haar\cite{zhang2004boosting}, and Dlib\cite{king2009dlib}. Therefore, considering the balance of accuracy and speed, the fast-RCNN algorithm is finally selected to achieve the ticket segmentation function. The example of one image contains multiple ticket regions is shown in Fig.\ref{fig3}.

\vspace{-2mm}
\begin{table}[!hbt]
\caption{These $900\times2048$ images are tested on Tesla V100 with 32G RAM.}
\begin{center}
 \begin{tabular}{ccc}
\hline
  \bfseries Algorithms & \bfseries AP50 & \bfseries FPS\\
\hline
Faster-RCNN & \textbf{0.94} & 100\\
SDD & 0.89 & 83\\
OpenCV-HAAR & 0.65 & 128\\
DLIB & 0.68 & \textbf{135}\\
\hline
\end{tabular}
\end{center}
\vskip -0.2in
\label{table2}
\end{table}

\subsubsection{Direction correction}
The convolution neural network is used to classify the tickets with different rotation directions, and the direction correction of tickets can be readily achieved. The specific steps are introduced in Algorithm\ref{algo:Direction}

\begin{algorithm}
\caption{Direction correction}
\begin{enumerate}[step a]
\item  Set $n_{class}$ as the total number of categories and $\theta$ as the angle identification degree of tickets, and they satisfy $\theta = \frac{360^{\circ}}{n_{class}}$
\item  Select the ticket sample set with rotation direction of 0 degree, and clockwise rotate the sample set $n\times\theta$ to form the training set data of each angle category when $n=0,1,2...n_{class-1}$ is taken.
\item  Hough transform is used to detect and enhance the linear features of samples in the training data set.
\item  The training data set is used to train the convolutional neural network classifier.
\item  The classifier is used to classify the samples to be identified, and the class $simple_{class}$ is obtained. The sample $\phi^{\circ}$ is rotated clockwise to complete the direction correction. where $\phi^{\circ}=(n_{class}-simple_{class})\times\theta$
\end{enumerate}
\label{algo:Direction}
\end{algorithm}

\begin{figure}[htbp]
%\vskip 0.2in
\begin{center}
\centerline{\includegraphics[width=1\columnwidth]{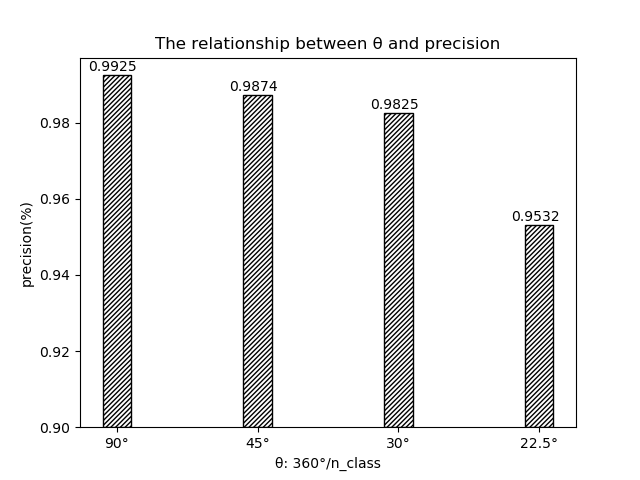}}
\caption{These are the results of testing the direction correction module when the images are rotated by $90^{\circ}$, $45^{\circ}$, $30^{\circ}$, and $22.5^{\circ}$.}
\label{fig4}
\end{center}
%\vskip -0.2in
\end{figure}

In step b, the $0^{\circ}$ ticket sample set should cover business ticket types and ticket styles as much as possible, so as to improve the generalization of direction correction classifier. At present, a variety of convolution neural network classifiers can complete this task. In order to fully measure the accuracy,speed and size of the classifier, we tried four models, Densenet-121, RESNeST-101\cite{zhang2020resnest}, EfficientNet-b7\cite{tan2019efficientnet} and DFL-CNN\cite{wang2018learning}. The results are shown in Table \ref{table3}. Finally, DFL-CNN is selected.The relationship between the angle identification degree $\theta$ and the accuracy $precision$ in the classifier is shown in Fig.\ref{fig4}. According to the specific needs of the business, the relationship between the accuracy rate and the angle identification of tickets can be balanced by properly setting $\theta$.

\vspace{-2mm}
\begin{table}[!hbt]
\caption{$\theta$=45°,nclass=8.The training data set was 8754 ticket images and the test data set is 800 ticket images,test on Tesla V100 with 32G RAM.}
\begin{center}
 \begin{tabular}{cccc}
\hline
  \bfseries Algorithms & \bfseries Acc & \bfseries FPS & \bfseries Model Size(MB)\\
\hline
Densenet-121 & 0.9650 & \textbf{112} & 53.6\\
RESNeST-101 & 0.9850 & 106 & 210\\
EfficientNet-b7 & 0.9787 & 98 & 507\\
DFL-CNN & \textbf{0.9874} & 103 & \textbf{46.2}\\
\hline
\end{tabular}
\end{center}
\vskip -0.2in
\label{table3}
\end{table}

\subsection{Tickets recognition module}
The ticket recognition module is the core module of the financial ticket intelligent recognition system. The input of the module is the preprocessed financial ticket image, and the output is the structured financial processing information. The module is further divided into three parts: ticket classification, ticket detection and recognition, and text structuring.

\subsubsection{Ticket classification}
The essence of ticket classification is image classification. Image classification technology based on deep learning has developed rapidly in the past 10 years. Since AlexNet\cite{krizhevsky2012imagenet} was proposed in 2012, VGG\cite{simonyan2014very}, GoogleNet\cite{szegedy2015going}, ResNet\cite{he2016deep}, DenseNet\cite{huang2017densely} and EfficientNet\cite{tan2019efficientnet} have been proposed one after another, and constantly refresh the highest accuracy of image classification. We try to use these models to solve the problem of ticket classification, among which EfficientNet-b7 is the best high precision XX\%. However, this accuracy is still insufficient for financial work. Therefore, we further study the overall characteristics of financial tickets data.In practical application, there are many kinds of tickets, which have the following two characteristics: A) There are similar tickets in different types of tickets, such as the receipt and payment receipt of the same bank. B) The same kind of ticket has great difference, such as different cities have different styles of toll tickets, but they are of the same category.Through the above characteristics, we draw a conclusion: in order to achieve higher classification accuracy, we need to pay more attention to the details of ticket image. Therefore, we use the idea of fine-grained image classification to solve the problem of ticket classification.

Fine grained image classification is a challenging research topic in the field of computer vision. Its goal is to identify the subclasses, which is consistent with the characteristics of ticket data.Due to the subtle differences between subcategories and large intra class differences, traditional classification algorithms have to rely on a large number of manual annotation information. In recent years, with the development of deep learning, deep convolution neural network brings new opportunities for fine-grained image classification.A large number of deep convolution based feature algorithms have been proposed, which promotes the rapid development of this field. At present, the fine-grained classification algorithm can be divided into two types: strong supervision and weak supervision. The so-called fine-grained image classification algorithm refers to the manual annotation of additional information such as label of label box and local area in model training. Some algorithms only use annotation information in model training, but not in image classification. This improves the practicability of the algorithm to a certain extent, but there is still a certain gap compared with the weakly supervised classification algorithm which only relies on the category label. Part-based R-CNNs\cite{zhang2014part} and pose normalized CNN\cite{branson2014bird} belong to this kind of algorithm.Weak supervised fine-grained classification algorithm only relies on category labels to complete classification, which is the mainstream algorithm in the field of fine-grained classification in recent years. This kind of algorithm uses attention mechanism and can achieve good classification performance without manual annotation information. B-CNN\cite{lin2015bilinear} and DFL-CNN are the main representatives of this kind of algorithm. Therefore, this system proposes Financial Ticket Classification (FTC) network on the basis of DFL-CNN, which is our early work. The details can be understood in detail in the paper \cite{article}. The accuracy of the network in ticket image classification reaches 99.36\%. At the same time, it shows a high classification confidence, which creates a strong condition for setting a higher confidence threshold and reducing the overall error rate of the system.

\begin{figure}[htbp]
%\vskip 0.2in
\begin{center}
\centerline{\includegraphics[width=1\columnwidth]{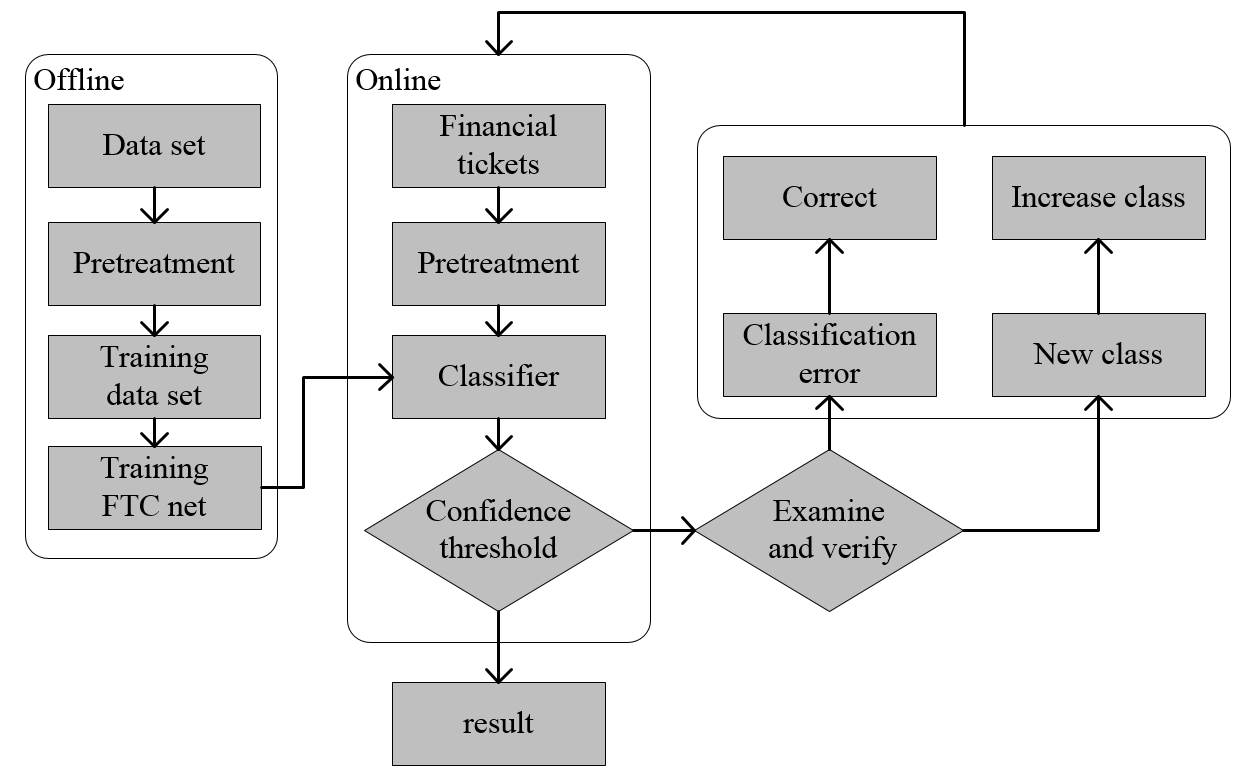}}
\caption{The framework of financial ticket classification algorithm}
\label{fig6}
\end{center}
%\vskip -0.2in
\end{figure}

At present, the classification model is trained offline. After training, it is deployed to the online financial ticket intelligent recognition system. The online business tickets pass through the classifier, and when the confidence reaches the threshold, the classification results are given, which will be pushed to the detection and recognition module. If the confidence level does not reach the threshold value, it will be reviewed manually and the classification results will be given by ticket auditors, then it will be fed into the detection and recognition module. As for these unfamiliar data and scarce data, they will be distinguished and pushed to the model training data set by the ticket warehouse to optimize the classification model. In the process of subsequent detection and recognition, the classification results are verified by ticket keyword information to complete the second cross-validation. When the classification error is found, these images will be returned to the manual audit step.

\subsubsection{Information detection and recognition}
The text detection and recognition algorithm based on deep learning has strong anti noise ability and robustness, and can detect and recognize characters in complex background and noise. The basic networks originated from image classification, detection, semantic segmentation and other visual processing tasks, such as VGG, ResNet, InceptionNet, DenseNet, Inside-Outside Net\cite{bell2016inside}, Se-Net\cite{hu2018squeeze}, etc., are used to extract the feature vectors of text regions in images.At the same time, many network frameworks originated from object detection and semantic segmentation tasks, such as SSD, YOLO, Faster-RCNN and so on, have been transformed to improve the accuracy and speed of image and text recognition tasks.At present, deep learning text detection and recognition algorithm achieves 92.4\% and 80\% accuracy on ICDAR 2013 and ICDAR 2015 data sets respectively\cite{zhu2016scene}.However, the accuracy of financial ticket recognition needs to be improved. We still analyze the specific characteristics of financial ticket data. Compared with the text image of natural scene, the particularity of ticket data includes:A)The background is simple. The background of the ticket is paper, and there are some interferences such as color, watermark or table line. However, compared with the open background in natural scene, the complexity is relatively low.B)There are few types of fonts. Due to the strong standardization of tickets, their fonts are concentrated in a few categories, and most of the Chinese tickets are in Song typeface, bold, Imitation Song typeface and Regular script. C) The shape of the text is standard. Most of the Chinese notes are horizontal straight line typesetting, a few of them are vertical straight line typesetting, no one is bending, twisting and oblique line typesetting.The above features make the difficulty of ticket detection and recognition relatively reduced, which can reduce some processing steps for complex background, font and special character shape in detection and recognition technology, which should be considered when improving the speed of detection and recognition algorithm.But at the same time, there are some special difficulties in ticket recognition.It mainly includes:A)Higher precision is required. Due to the particularity of financial work, the accuracy requirement of ticket detection and recognition is much higher than that of scene character recognition.B)There are special noises. The ticket wrinkle, wear and the overlapping and dislocation of characters in printing will cause the ticket text to lose the obvious boundary with the background, and there will be some problems such as text deformation, ambiguity and incompleteness. These problems are rarely seen in the text of natural scenes.At the same time, when the quality of photo taking and scanning is not high, the problems of shadow occlusion, uneven illumination, perspective transformation and text bending will also be caused.C)There are more text areas. Compared with the natural scene text image, the ticket text is arranged densely, and the number of text areas is usually one order of magnitude more, which puts forward higher requirements for detection and recognition efficiency.

\begin{figure}[htbp]
%\vskip 0.2in
\begin{center}
\centerline{\includegraphics[width=1\columnwidth]{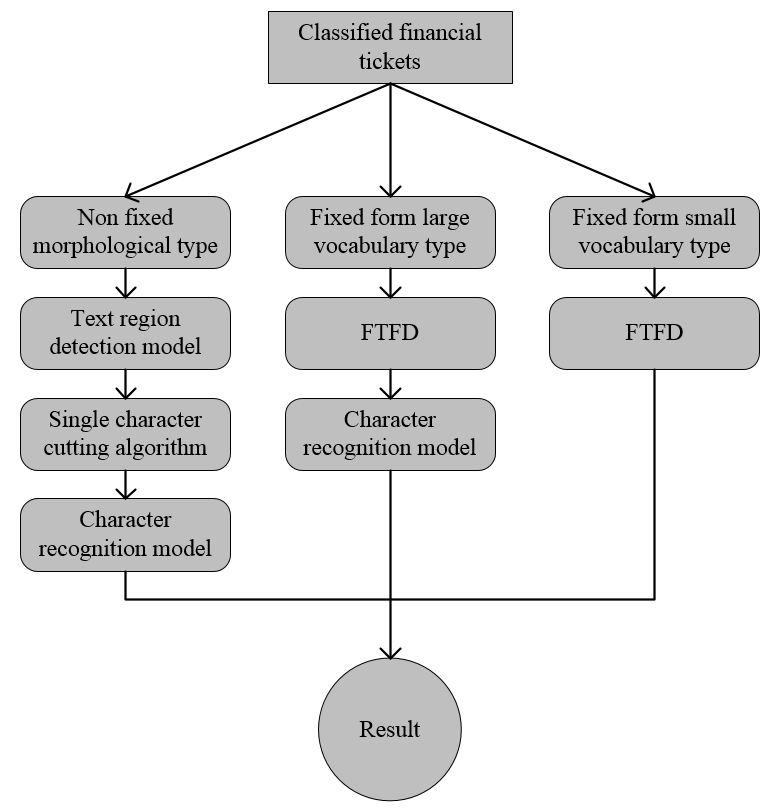}}
\caption{The framework of detection and recognition module}
\label{fig7}
\end{center}
%\vskip -0.2in
\end{figure}

Financial ticket detection and recognition is the key link of our system. In order to improve the efficiency of detection and recognition, we fully analyze the semantic features of financial ticket data, and treat different types of tickets according to the simple and effective principle to form the algorithm in this paper.Compared with the text image of the natural scene, the financial ticket image has more small and clustered text areas. According to the statistics of 180000 tickets, on each ticket, there are 43 text areas could be detected on average, but only a small number of information areas are needed in financial work. In addition, some tickets need to be identified with fixed regional shape, and the recognition content is less than 100 small vocabulary targets. Therefore, the detection and recognition algorithm designed according to the general text detection and recognition process of region detection, which consists by character segmentation, character recognition, and result structure, will lead to the following problems: A) The full surface detection and recognition will produce a large amount of redundant information for financial work, which will reduce the efficiency of system ticket recognition and waste computing resources. B) All single words of ticket character recognition are completed by the Chinese character recognition model, which increases the running burden of the server and reduces the recognition accuracy. Therefore, financial tickets are divided into three types: I) Fixed form small vocabulary type. The recognition content of this type of ticket is fixed form small vocabulary targets which usually include specific Chinese characters, English letters, and numbers. II) Fixed form large vocabulary type. The identification content of the ticket is a fixed form field, but it involves a large vocabulary of Chinese characters. III) Non-fixed form types. The identification contents of the ticket are non-fixed form fields, including Chinese characters, English letters, punctuation marks, and special symbols. As shown in Fig.\ref{fig7}, the system contains different detection and recognition processes following the characteristics of three types of tickets.

\begin{figure*}[htbp]
%\vskip 0.2in
\begin{center}
\centerline{\includegraphics[width=2\columnwidth, height=10cm]{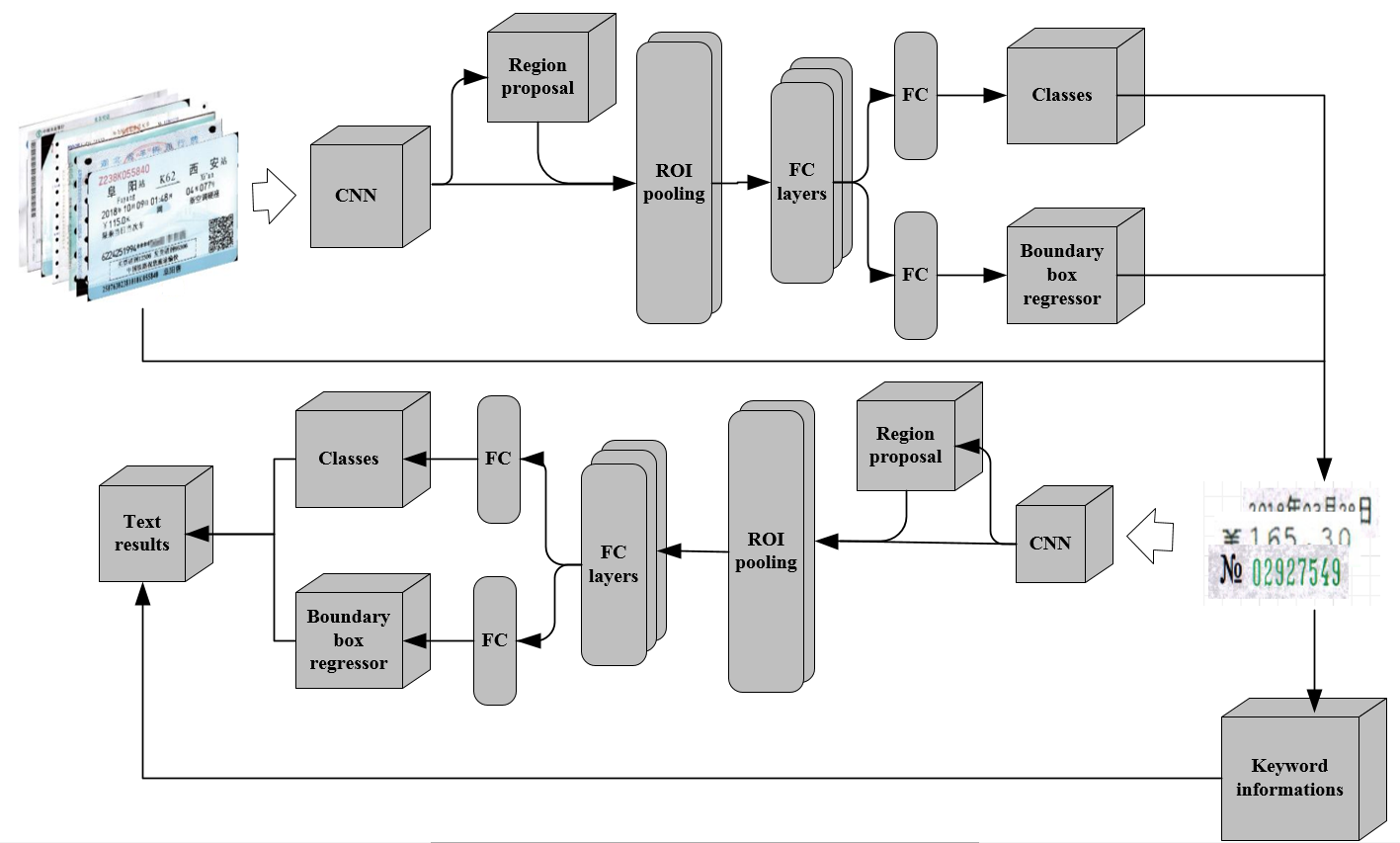}}
\caption{It is the structure of FTFDNet, which extracts the text region at the first stage, then recognizes the words at the second stage.}
\label{fig8}
\end{center}
%\vskip -0.2in
\end{figure*}

\begin{table*}[!hbt]
\caption{The time consumption of recognition task is not only impacted by resolution but the number of text regions, which causes the differences among quota tickets, train tickets, and taxi tickets.}
\begin{center}
 \begin{tabular}{cccc}
    \hline
    \multirow{2}*{types of tickets} & \multirow{2}*{average resolution} &
    \multicolumn{2}{c}{time consumption for one ticket(ms)}\\
    \cline{3-4} 
    & & FTFDNet & general algorithm\\ \hline
    VAT tickets & $1024\times2048$ & \textbf{88.67} & $844.33$\\\hline
    Quota tickets & $600\times1024$ & \textbf{55} & $153.67$\\\hline
    Train tickets & $600\times1024$ & \textbf{85.33} & $334.33$\\\hline
    Taxi tickets & $520\times1500$ & \textbf{72.67} & $380.67$\\\hline
\end{tabular}
\end{center}
\vskip -0.2in
\label{table4}
\end{table*}

\begin{figure*}[htbp]
%\vskip 0.2in
\begin{center}
\centerline{\includegraphics[width=2\columnwidth]{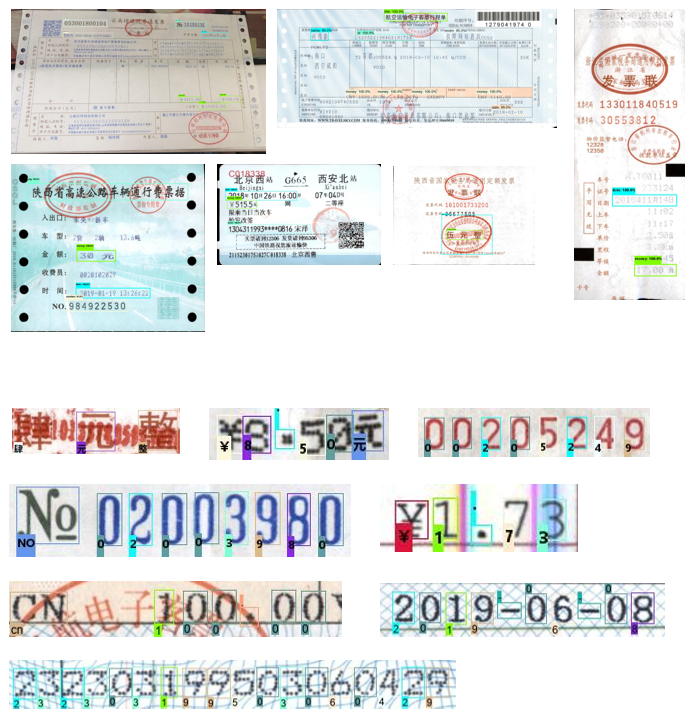}}
\caption{Some results of content recognition of type I and type II tickets using FTFDNet.}
\label{fig16}
\end{center}
%\vskip -0.2in
\end{figure*}

In general, the time consumption of a single sample in the ticket recognition system can be expressed by: 
\begin{equation}
    T = \alpha \times(w+h)+\beta \times A_{text}+ \gamma \times A_{information} + C
\end{equation}
Where $w+h$ is the resolution of the sample image, $A_{text}$ is the sum of the area of the image text region, $A_{information}$ is the sum of the area of each business information region in the image, $C$ is the wear constant, which represents the loss time of the system reading, transmission and structure, $\alpha, \beta, \gamma$ are the coefficients for each influencing factor. The formula shows that the time consumption of a single sample recognition is directly proportional to the sample image size, text region, and information region. In order to reduce the system time consumption, We divide the algorithm into three branches according to the types of tickets:

\textbf{Type I ticket detection and recognition.}
According to the characteristics of type I, based on Faster-RCNN, a FTFDNet is established.Faster-RCNN is a two stage target detection algorithm. The Region Proposal Network (RPN) is used to extract the target candidate regions from the feature map given by the backbone network, and then the uniform size ROI region feature is obtained by ROI pooling, which is sent to multiple classifiers for target classification and position regression. Faster-RCNN is a single, unified network for object detection,which has high detection accuracy and speed. 
As shown in Fig.\ref{fig8}, the FTFDNet firstly uses Faster-RCNN to detect the target area which needs to be identified in the whole ticket, and forms the keyword information of the final recognition result according to the target category and location information. At the same time, it continues to carry out single character target detection on the trimmed target area image, synthesizes the keyword information and detection results of FTFDNet to form the character recognition result. This approach only detects the interest region, which avoids redundant time consumption and computing source consumption. Mathematically, $\beta$ tends to $0$. Meanwhile, according to the characteristics of some tickets recognition targets with relatively fixed shape and small vocabulary, the classification information and position information in the process of FTFDNet are used to replace the character recognition model and the result structure process, which greatly reduces the $\gamma$ and $C$ and effectively improves the system recognition efficiency.Fig.\ref{fig16} shows some results.

\begin{figure}[htbp]
%\vskip 0.2in
\begin{center}
\centerline{\includegraphics[width=1\columnwidth]{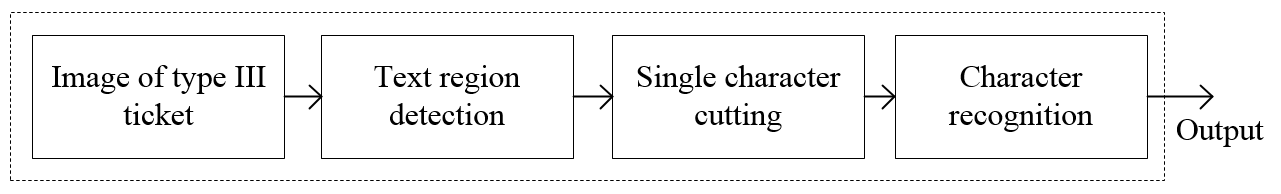}}
\caption{Algorithm framework for detecting and identifying type III tickets}
\label{fig12}
\end{center}
%\vskip -0.2in
\end{figure}

\textbf{Type III ticket detection and recognition.}
For type III tickets, the financial business needs more information, and the same information expression forms of different types of tickets are quite different.Therefore, full face character recognition is adopted to process type III tickets. The specific algorithm framework is shown in Fig.\ref{fig12}. First of all, all text areas on the ticket surface are detected. CTPN, EAST, SegLink, TextBoxes, PixelLink and other deep learning models can complete this task well, and each has its own characteristics.CTPN uses Bi-LSTM module to extract the context features of the image where the characters are located to improve the recognition accuracy of text blocks;EAST model supports quadrilateral detection in any direction;SegLink model can predict single small text blocks, then link them into words, and predict slanted text;TextBoxes adjusts the size of anchor box and convolution kernel to rectangle, so as to adapt to the characteristics of slender text;Pixellink does not use conventional regression method, but uses instance segmentation method to predict text lines.Considering that most of the Chinese bills are horizontally aligned text lines, CTPN is selected as the text area detection model,The detection effect is shown in Fig.\ref{fig9}.

\begin{figure}[htbp]%\vskip 0.2in
\begin{center}
\centerline{\includegraphics[width=1\columnwidth]{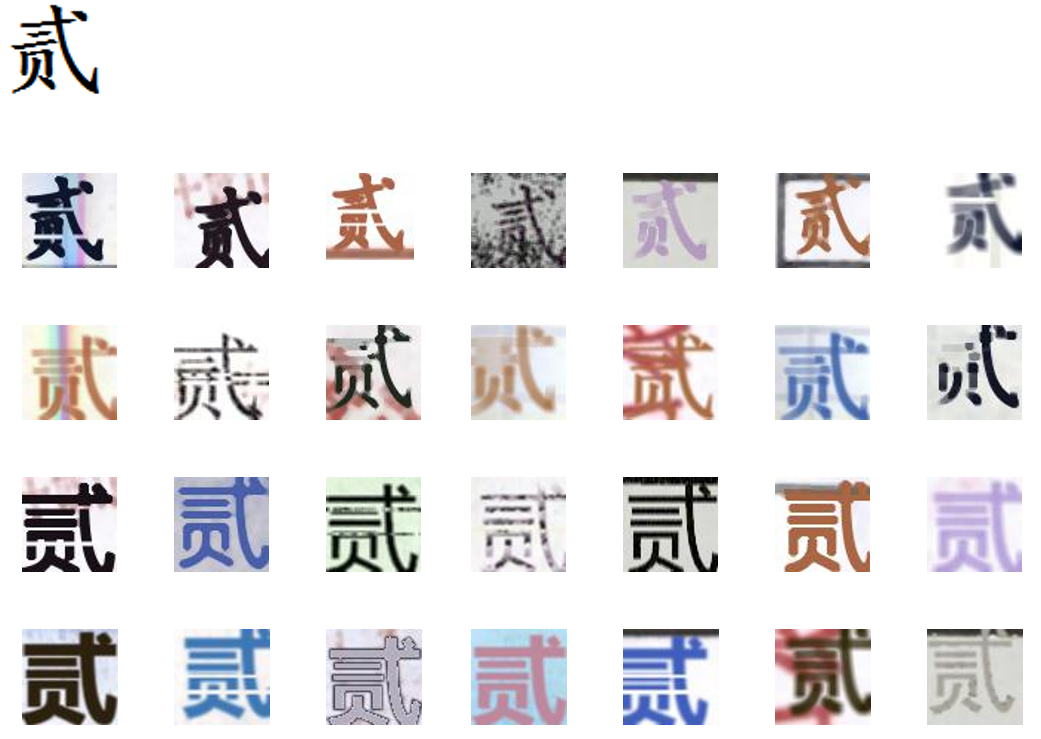}}
\caption{For the partial generated single character image data, the feature distribution of the generated character image must be consistent with the underlying feature distribution of the application scene data image as far as possible.}
\label{fig14}
\end{center}
%\vskip -0.2in
\end{figure}

After getting the image of each text region, the next step is to recognize the characters in the region image. For this task, two types of algorithms can be selected.Algorithm 1: end to end character recognition algorithm represented by CRNN, which uses CNN to extract image features, uses RNN to fuse feature vectors to extract context features of character sequences, and obtains the probability distribution of each column feature. Finally, the text sequence is predicted by CTC.The advantage of this algorithm is to support end-to-end joint training, strong noise resistance and strong robustness. But it needs a lot of manual annotation data for training. It is not easy to get a lot of data;Algorithm 2: the task of image recognition in text area is divided into two parts.In the first segment, a single character image is cut out, and in the second segment, image classification is used to complete character recognition.Compared with algorithm 1, the ability of algorithm 2 to resist noise is obviously weaker. The main reason is that algorithm 2 treats each character prediction as an independent event, without considering the context between characters.In algorithm 1, the RNN is used to extract the context features of character sequences. Therefore, when a character is noisy and difficult to recognize, it can still give a relatively correct prediction through the context.The advantage of algorithm 2 is that it is relatively easy to generate data sets. With algorithm 2, engineering projects can usually have a quick start. Algorithm 2 is used in our system.For character segmentation task, there are many mature algorithms, such as connected region algorithm, projection histogram algorithm, water drop algorithm and so on. When the text area in the ticket image are clear and standard, these algorithms can achieve good results. But when the text area in the ticket image have adhesion, overlap or complex noise, it is difficult to use these algorithms.Using the object detection model based on deep learning to take characters as the target, and detect them from the image of text area is an effective method to solve the text adhesion, overlapping and complex noise. In this paper, Faster-RCNN is used to realize character cutting.The feature distribution of the data set constructed must be consistent with the underlying feature distribution of the application scene data image as far as possible.The feature distribution of the data set constructed must be consistent with the underlying feature distribution of the application scene data image to the greatest extent.Therefore, we extract 482 kinds of tickets background texture, character color, font and other features, as well as table line interference, uneven illumination, wrinkles and wear and other common noise to form a 7-D data construction space. Using 4088 common Chinese characters of financial tickets, 6.77 million pieces of data are generated to form the character recognition model training data set.Fig.\ref{fig14} is a partial example of a single character image generated data.On this basis, a variety of CNN classifiers are verified. Finally, DFL-CNN is selected as the character recognition classifier, with an accuracy of 99.28\%.

\begin{figure}[htbp]%\vskip 0.2in
\begin{center}
\centerline{\includegraphics[width=1\columnwidth]{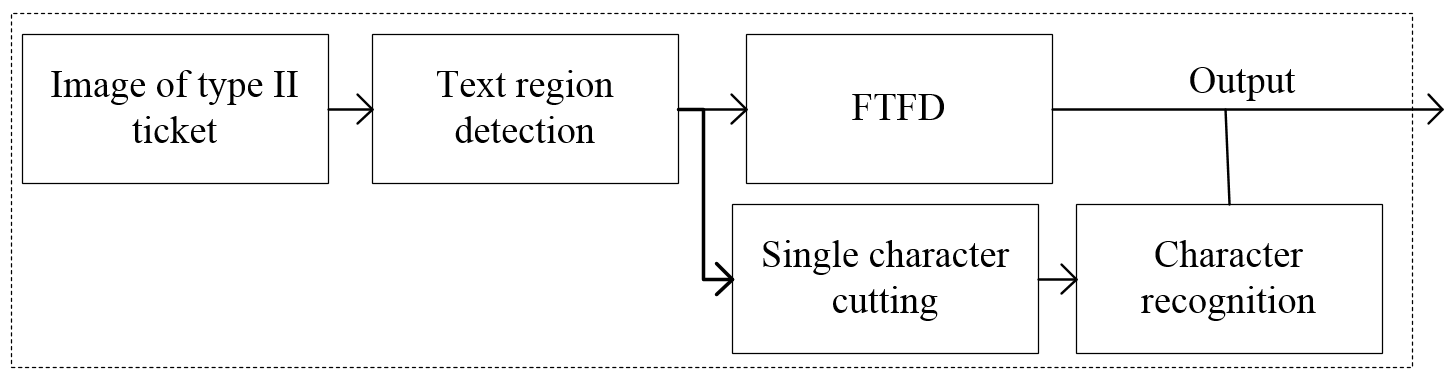}}
\caption{Algorithm framework for detecting and identifying type II tickets.}
\label{fig13}
\end{center}
%\vskip -0.2in
\end{figure}

\begin{figure}[htbp]%\vskip 0.2in
\begin{center}
\centerline{\includegraphics[width=1\columnwidth]{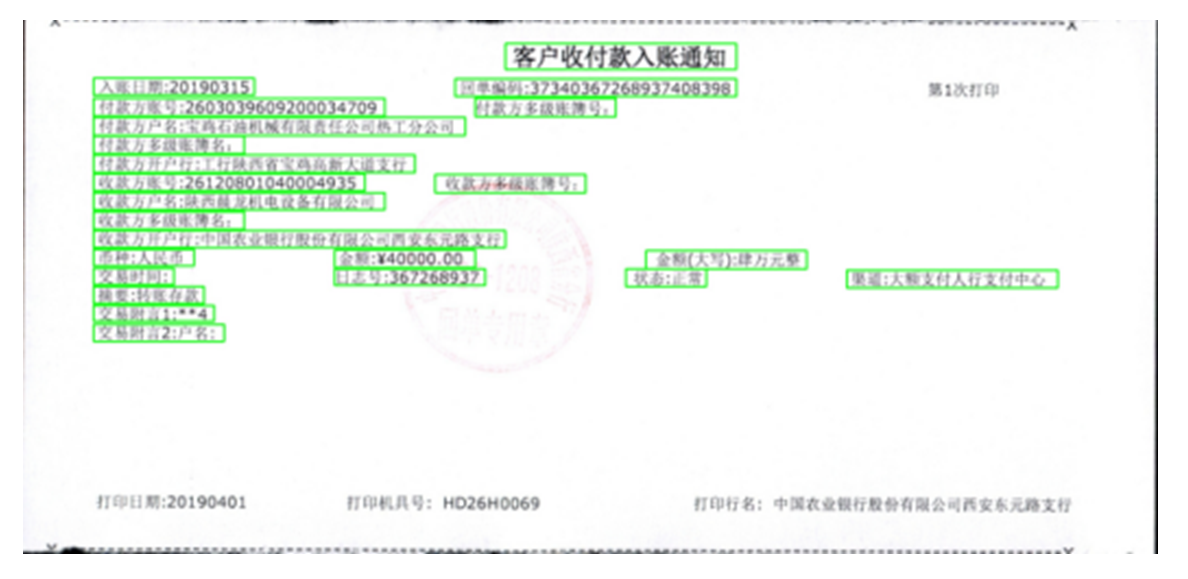}}
\caption{It is an exemplar of full surface detection of bank notes.}
\label{fig9}
\end{center}
%\vskip -0.2in
\end{figure}

\textbf{Type II ticket detection and recognition.}
Type II ticket is the ticket with name field in the text recognition area of plane trip ticket and train ticket, which has the characteristics of type I ticket and type III ticket. The recognition algorithm framework is shown in Fig.\ref{fig13}. The non name area adopts the same FTFDNet network as type I ticket, and the name area adopts the same segmented recognition algorithm as type III ticket.

The FTFDNet significantly improves the recognition speed of value-added tax(VAT) tickets, quota tickets, toll tickets, taxi tickets, plane tickets, etc. The results of the speed comparison based on a 2000 tickets dataset are shown in Table \ref{table4}.

Among them, the general algorithm is implemented according to the process of text region detection, character segmentation, character recognition, and information structure. It uses CTPN\cite{tian2016detecting} to detect the text area, Faster-RCNN to complete character segmentation, and DFL-CNN to realize character recognition.

Based on the statistical analysis of 716872 tickets generated by 276 companies in 2019, type-I tickets accounted for 65.69\%, type-II tickets accounted for 2.58\%, and type-III tickets accounted for 31.73\%. According to the distribution of ticket types, the average speed of ticket recognition is increased by 3.88 times after using FTFDNet.

\subsubsection{Text structuring}
The text structuring of the financial ticket is to convert the identified disordered texts into formatted texts according to the needs of the financial business. The purpose is to extract the information fields needed by financial work. Taking bank receipts as an example, text structuring is to extract the information, such as ``ticket title, transaction date, amount, currency, purpose, remarks, postscript, serial number, name of payer and payee, account number, bank name'', from the whole identification result for accounting entry, which is an indispensable link in financial ticket identification. In the detection and recognition algorithm proposed in this paper, the type, content, and location information of the detection target have been given synchronously in the process of type-I and type-II tickets orientational detection. Therefore, there is no need for text structuring, which is an important reason why the algorithm can improve the speed. After the recognition of type-III tickets, the disordered text with location information needs to be extracted by a structured algorithm. The main difficulties are as follows: 1) The structure is not fixed, and there are many positional relationships between the keywords and their corresponding values. 2) The keywords may have more or less words due to recognition errors. 3) There are many kinds of tickets and the structure of them are complex. In view of the above problems, Algorithm\ref{algo:structuring} is designed for this.
\begin{algorithm}
\caption{text structuring}
\begin{algorithmic}[1]
\State
\textbf{Inputs:} $inputList$: text recognition result list with location information. $keyWordsList$: keyword list for structured fields. $ticketType$: type of input ticket.\\
\textbf{Outputs:} $resultDictionary$: A dictionary of structured result. $positionList$: A list that need to be matched with location information.\\
\textbf{Begin}
\If{ticketType==III}
    \For{char in inputList}
        \For{oneKey in keyWordsList}
            \If{char includes oneKey}
                \If{there are remains on left side}
                    \State put them into positionList
                    \State text structuring(rest)
                \ElsIf{there are remains on right side}
                    \If{right-side remains has keyword}
                        \State put oneKey into positionList
                        \State text structuring(rest)
                    \Else
                        \State resultDictionary['oneKey'] ← rest
                    \EndIf
                \Else
                    \State put oneKey into positionList
                \EndIf
            \EndIf
        \EndFor
    \EndFor
    \If{positionList is not empty}
        \State Matching key in positionList by location matching algorithm
    \EndIf
\EndIf

\end{algorithmic}
\label{algo:structuring}
\end{algorithm}

\subsection{Accounting entry module}
Accounting subject is a category of accounting element object's specific content classification accounting. Ticket accounting entry is to divide a ticket into corresponding accounts according to its information, so that the economic information in the original ticket voucher can be transformed into accounting language and is used as the direct basis for registering account books. Through identification, each VAT ticket, for instance, can obtain nine items of information: the name of the payee and payer, ticket code, number, verification code, type, abstract, transaction details, and ticket date. Among them, the transaction details include 4263 kinds of goods or service categories. According to these analyses, the accounting can divide the ticket into 34 subjects such as office expenses, travel expenses, labor expenses, etc. In order to realize this function, we take bill information as input and convert account entry into text classification. In the field of text classification, deep learning algorithm also has obvious advantages over traditional machine learning algorithm.Fast-Text\cite{wu1992fast}, Text-CNN\cite{kim2014convolutional}, TextRNN\cite{liu2016recurrent}, TextRCNN\cite{lai2015recurrent}, HAN\cite{yang2016hierarchical}, BERT\cite{devlin2018bert} and other models have been proposed one after another, which makes the accuracy of text classification improve continuously.Our system constructs the accounting entry module based on the pretrained BERT. Firstly, the labeled accounting entry data are sorted out and proportionally divided into the training data set, the testing set, and the verification set. Sequentially, using the data fine-tune BERT framework to get a classifier for accounting entry. The experimental result illustrates the accuracy of the classifier for ticket accounting entry can reach 96.5\%.

\begin{figure}[htbp]
%\vskip 0.2in
\begin{center}
\centerline{\includegraphics[width=0.3\columnwidth, height=6cm]{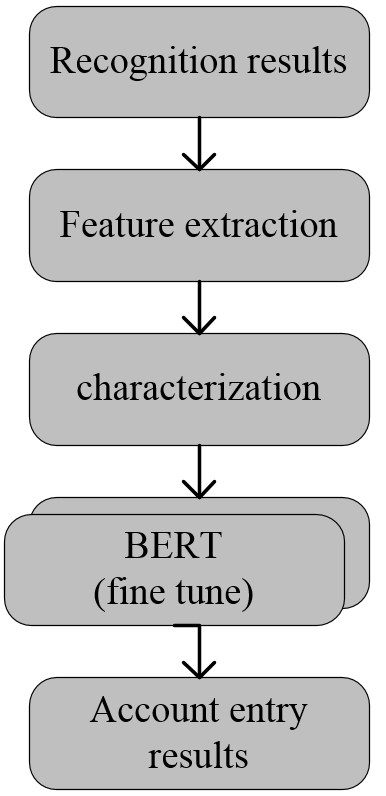}}
\caption{Accounting entry module}
\label{fig10}
\end{center}
%\vskip -0.2in
\end{figure}

\section{Experiment}
In order to accurately monitor the recognition accuracy, operation speed, and operating conditions of the financial ticket intelligent recognition system without human intervention, the time consumption and recognition accuracy of the ticket recognition module were tested by using common types of tickets, and the computing source consumption of our system was observed.

\subsection{Data}
Through long-term collation, we divide financial tickets into 12 major-classes with a total of 482 sub-classes. Our financial ticket intelligent recognition system has accumulated more than 10 million real tickets which continue to increase by approximately 10,000 per day. The names of the major classes of tickets and the number of corresponding sub-classes are shown in Table\ref{table5}. 

\begin{table}[!h]
 \renewcommand{\arraystretch}{1.3}
 \caption{This table shows the specific name for 12 major-classes and the number of sub-classes that each main-class contains.}
 \label{table5}
 \centering
 \begin{tabular}{ccc}
  \hline
  \bfseries Class Index & \bfseries Class name & \bfseries Number of sub-classes \\
  \hline
  1 & Bank tickets & 149 \\
  2 & VAT tickets & 104 \\
  3 & Toll tickets & 85 \\
  4 & Quota tickets & 51 \\
  5 & Admission tickets & 25 \\
  6 & Common printed tickets & 21 \\
  7 & Taxi tickets & 21 \\
  8 & Plane tickets & 1 \\
  9 & Train tickets & 1 \\
  10 & Insurance policy tickets & 7 \\
  11 & Receipts & 2 \\
  12 & Others & 15 \\
  \hline
 \end{tabular}
\end{table}

\begin{table*}[!h]
 \renewcommand{\arraystretch}{1.3}
 \caption{We pick up 200 images for each six types of images in the table as our testing dataset. It is noticed that there is a big difference between the number of text region and the number of information region, which is where we are inspired from for FTFDNet.}
 \label{table6}
 \centering
 \begin{tabular}{ccccc}
  \hline
  \bfseries Type of tickets & \bfseries average resolution & \bfseries the number of text region & \bfseries the number of information region & \bfseries Amount of tickets \\
  \hline
  VAT tickets& $2047\times1210$ & $75$ & $7$ & $200$ \\
  Toll tickets& $1057\times904$ & $18$ & $5$ & $200$ \\
  Quota tickets & $1283\times894$ & $16$ & $1$ & $200$ \\
  Train tickets & $534\times1530$ & $17$ & $3$ & $200$ \\
  Taxi tickets & $1047\times667$ & $52$ & $2$ & $200$ \\
  Bank receipt & $2420\times1733$ & $39$ & $11$ & $200$\\
  \hline
 \end{tabular}
\end{table*}

A total of 1200 financial tickets including VAT tickets, toll tickets, quota tickets, train tickets, taxi tickets, and domestic payment receipts of Bank of China are sorted out. See Table \ref{table6} for the ticket distribution, where the number of text areas is the average number of all detectable text areas on the ticket surface, and the number of information areas is the number of information text areas required by financial business.

\subsection{Implementation}

The financial ticket intelligent recognition system is built by TensorFlow and Pytorch framework. The system deployment server operating system is CentOS Linux release (core) 7. We use 2 pieces of Tesla V100 card with 32G video memory each, and the CPU frequency is 2.20GHz.

\subsection{Results}
\subsubsection{Time consumption}
According to the classification statistics, the No.1-3 are type-I ticket, No.4 is type-II ticket, and No.5 is type-III ticket. The results show the average single sample identification time is 175.67ms, which can meet the real-time requirements of financial business. 

\begin{table}[!htb]
 \renewcommand{\arraystretch}{1.3}
 \caption{This table shows the overall time consumption, which includes tickets preprocessing module and tickets recognition module.}
 \label{table7}
 \centering
 \begin{tabular}{ccc}
  \hline
  \bfseries Class Index & \bfseries Class name & \bfseries time(ms) \\
  \hline
  1 & VAT tickets& 88.67 \\
  2 & Quota tickets & 55 \\
  3 & Taxi tickets & 72.67 \\
  4 & Train tickets & 85.33 \\
  5 & Bank receipt & 678.33 \\
  \hline
 \end{tabular}
\end{table}

\subsubsection{Accuracy}
The recognition accuracy of the system is measured by the correct rate of string area recognition and the correct rate of whole ticket recognition. The correct rate of string recognition is represented by $P_{char}$, which is the proportion of the total number of strings whose characters are correctly recognized. The calculation formula is:
\begin{equation}
    P_{char} = \frac{\sum R_{char}}{\sum N_{char}}
\end{equation}
where $R_{char}$ is the correctly recognized string in a single sample, and $N_{char}$ is the total number of strings in a single sample. The correct rate of the whole ticket identification is the proportion of the total amount of financial tickets whose business information fields have been correctly identified. The calculation follows:
\begin{equation}
    P_{ticket} = \frac{R_{ticket}}{N_{ticket}}
\end{equation}
where $R_{ticket}$ is the number of financial tickets whose business information fields have been correctly identified, and $N_{ticket}$ is the total number of tickets.

\begin{table}[!htb]
 \renewcommand{\arraystretch}{1.3}
 \caption{It gives the results of recognition accuracy for both strings and whole ticket. AccS is the recognition accuracy of string region. AccW is the recognition accuracy of whole ticket.}
 \label{table8}
 \centering
 \begin{tabular}{cccc}
  \hline
  \bfseries Class Index & \bfseries Class name & \bfseries AccS & \bfseries AccW \\
  \hline
  1 & VAT tickets & $96.12\%$ & $87\%$ \\
  2 & Quota tickets & $99.5\%$ & $99.5\%$ \\
  3 & Taxi tickets & $99\%$ & $97.5\%$ \\
  4 & Train tickets & $96\%$ & $87\%$ \\
  5 & Bank receipt & $94.75\%$ & $87\%$ \\
  \hline
 \end{tabular}
\end{table}

The experimental result illustrates that the average recognition accuracy of the six kinds of tickets is 97.07\%, and the average recognition accuracy of the whole ticket is 91.6\%. Under the condition of a reasonable threshold setting, the system can reduce the labor cost by more than 70\%. Moreover, it can meet the requirements of financial work with low fault tolerance.

\subsubsection{Computing source consumption}
The financial ticket intelligent recognition system designed in this paper is composed of 21 deep learning models. type-I tickets and type-II tickets are deployed as the same service, while type-III tickets are deployed separately, accounting for the server computing resources as shown in the Table\ref{table9}.
\begin{table}[!htb]
 \renewcommand{\arraystretch}{1.3}
 \caption{Computing source consumption, where MaxM stands for the maximum memory occupation and MinM stands for the minimum memory occupation.}
 \label{table9}
 \centering
 \begin{tabular}{cccc}
  \hline
  \bfseries Recognition service & \bfseries batch & \bfseries MaxM & \bfseries MinM \\
  \hline
  Recognition for bank tickets & 20 & 9.5G & 5.2G\\
  Others & 5 & 7.8G & 4.1G\\
  \hline
 \end{tabular}
\end{table}

\section{Conclusion}

\begin{figure*}[htbp]
%\vskip 0.2in
\begin{center}
\centerline{\includegraphics[width=2\columnwidth]{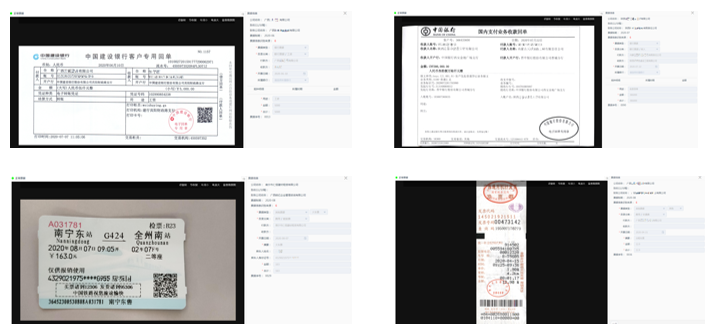}}
\caption{The system interface.}
\label{fig17}
\end{center}
%\vskip -0.2in
\end{figure*}

According to the structure of four modules and two branches, the intelligent financial ticket recognition system is constructed, and it is applied to specific business. FTFDNet is designed for the recognition task of the tickets with fixed target shape, which could effectively improve the recognition speed and enhance the practicability of the system. From the experimental data observation, the average recognition accuracy of the ticket information area of the system is 97.07\%, and the average time consumption for single ticket is 175.67ms. Therefore, this system satisfies both low fault tolerance and real-time processing requirements. Moreover, its practical value has been tested in online business, which makes a beneficial attempt for the in-depth application of deep learning in the finance and taxation industry.The system interface is shown in the Fig.\ref{fig17}.In addition, the ticket warehouse is proposed and implemented, which could achieve the automatic iterative optimization to enhance the power of the system and benefit the future business.

\bibliography{references}

\begin{thebibliography}{10}
\providecommand{\url}[1]{#1}
\csname url@samestyle\endcsname
\providecommand{\newblock}{\relax}
\providecommand{\bibinfo}[2]{#2}
\providecommand{\BIBentrySTDinterwordspacing}{\spaceskip=0pt\relax}
\providecommand{\BIBentryALTinterwordstretchfactor}{4}
\providecommand{\BIBentryALTinterwordspacing}{\spaceskip=\fontdimen2\font plus
\BIBentryALTinterwordstretchfactor\fontdimen3\font minus
  \fontdimen4\font\relax}
\providecommand{\BIBforeignlanguage}[2]{{%
\expandafter\ifx\csname l@#1\endcsname\relax
\typeout{** WARNING: IEEEtran.bst: No hyphenation pattern has been}%
\typeout{** loaded for the language `#1'. Using the pattern for}%
\typeout{** the default language instead.}%
\else
\language=\csname l@#1\endcsname
\fi
#2}}
\providecommand{\BIBdecl}{\relax}
\BIBdecl

\bibitem{miikkulainen2019artificial}
R.~Miikkulainen, J.~Liang, E.~Meyerson, A.~Rawal, D.~Fink, O.~Francon, B.~Raju,
  H.~Shahrzad, A.~Navruzyan, N.~Duffy \emph{et~al.}, ``Artificial intelligence
  in the age of neural networks and brain computing,'' 2019.

\bibitem{feng2019computer}
X.~Feng, Y.~Jiang, X.~Yang, M.~Du, and X.~Li, ``Computer vision algorithms and
  hardware implementations: A survey,'' \emph{Integration}, vol.~69, pp.
  309--320, 2019.

\bibitem{charniak2019introduction}
E.~Charniak, \emph{Introduction to deep learning}.\hskip 1em plus 0.5em minus
  0.4em\relax The MIT Press, 2019.

\bibitem{solis2019domain}
A.~I. Solis and P.~Nava, ``Domain specific architectures, hardware acceleration
  for machine/deep learning,'' in \emph{Disruptive Technologies in Information
  Sciences II}, vol. 11013.\hskip 1em plus 0.5em minus 0.4em\relax
  International Society for Optics and Photonics, 2019, p. 1101307.

\bibitem{o2019deep}
N.~O’Mahony, S.~Campbell, A.~Carvalho, S.~Harapanahalli, G.~V. Hernandez,
  L.~Krpalkova, D.~Riordan, and J.~Walsh, ``Deep learning vs. traditional
  computer vision,'' in \emph{Science and Information Conference}.\hskip 1em
  plus 0.5em minus 0.4em\relax Springer, 2019, pp. 128--144.

\bibitem{jha2019automation}
M.~Jha, M.~Kabra, S.~Jobanputra, and R.~Sawant, ``Automation of cheque
  transaction using deep learning and optical character recognition,'' in
  \emph{2019 International Conference on Smart Systems and Inventive Technology
  (ICSSIT)}.\hskip 1em plus 0.5em minus 0.4em\relax IEEE, 2019, pp. 309--312.

\bibitem{srivastava2019optical}
S.~Srivastava, J.~Priyadarshini, S.~Gopal, S.~Gupta, and H.~S. Dayal, ``Optical
  character recognition on bank cheques using 2d convolution neural network,''
  in \emph{Applications of Artificial Intelligence Techniques in
  Engineering}.\hskip 1em plus 0.5em minus 0.4em\relax Springer, 2019, pp.
  589--596.

\bibitem{ha2017recognition}
H.~T. Ha, ``Recognition of invoices from scanned documents.'' in \emph{RASLAN},
  2017, pp. 71--78.

\bibitem{shreya2019optical}
S.~Shreya, Y.~Upadhyay, M.~Manchanda, R.~Vohra, and G.~D. Singh, ``Optical
  character recognition using convolutional neural network,'' in \emph{2019 6th
  International Conference on Computing for Sustainable Global Development
  (INDIACom)}.\hskip 1em plus 0.5em minus 0.4em\relax IEEE, 2019, pp. 55--59.

\bibitem{sun2019template}
Y.~Sun, X.~Mao, S.~Hong, W.~Xu, and G.~Gui, ``Template matching-based method
  for intelligent invoice information identification,'' \emph{IEEE Access},
  vol.~7, pp. 28\,392--28\,401, 2019.

\bibitem{zhang2019research}
J.~Zhang, F.~Ren, H.~Ni, Z.~Zhang, and K.~Wang, ``Research on information
  recognition of vat invoice based on computer vision,'' in \emph{2019 IEEE 6th
  International Conference on Cloud Computing and Intelligence Systems
  (CCIS)}.\hskip 1em plus 0.5em minus 0.4em\relax IEEE, 2019, pp. 126--130.

\bibitem{cesarini2003analysis}
F.~Cesarini, E.~Francesconi, M.~Gori, and G.~Soda, ``Analysis and understanding
  of multi-class invoices,'' \emph{Document Analysis and Recognition}, vol.~6,
  no.~2, pp. 102--114, 2003.

\bibitem{palm2017cloudscan}
R.~B. Palm, O.~Winther, and F.~Laws, ``Cloudscan-a configuration-free invoice
  analysis system using recurrent neural networks,'' in \emph{2017 14th IAPR
  International Conference on Document Analysis and Recognition (ICDAR)},
  vol.~1.\hskip 1em plus 0.5em minus 0.4em\relax IEEE, 2017, pp. 406--413.

\bibitem{blanchard2019automatic}
J.~Blanchard, Y.~Bela{\"\i}d, and A.~Bela{\"\i}d, ``Automatic generation of a
  custom corpora for invoice analysis and recognition,'' in \emph{2019
  International Conference on Document Analysis and Recognition Workshops
  (ICDARW)}, vol.~7.\hskip 1em plus 0.5em minus 0.4em\relax IEEE, 2019, pp.
  1--1.

\bibitem{yi2019dual}
F.~Yi, Y.-F. Zhao, G.-Q. Sheng, K.~Xie, C.~Wen, X.-G. Tang, and X.~Qi, ``Dual
  model medical invoices recognition,'' \emph{Sensors}, vol.~19, no.~20, p.
  4370, 2019.

\bibitem{klein2004results}
B.~Klein, S.~Agne, and A.~Dengel, ``Results of a study on invoice-reading
  systems in germany,'' in \emph{International workshop on document analysis
  systems}.\hskip 1em plus 0.5em minus 0.4em\relax Springer, 2004, pp.
  451--462.

\bibitem{kieri2012context}
A.~Kieri, ``Context dependent thresholding and filter selection for optical
  character recognition,'' 2012.

\bibitem{bailey2007single}
D.~G. Bailey and C.~T. Johnston, ``Single pass connected components analysis,''
  in \emph{Proceedings of image and vision computing New Zealand}, 2007, pp.
  282--287.

\bibitem{sauvola2000adaptive}
J.~Sauvola and M.~Pietik{\"a}inen, ``Adaptive document image binarization,''
  \emph{Pattern recognition}, vol.~33, no.~2, pp. 225--236, 2000.

\bibitem{platt1998sequential}
J.~Platt, ``Sequential minimal optimization: A fast algorithm for training
  support vector machines,'' 1998.

\bibitem{ratsch2001soft}
G.~R{\"a}tsch, T.~Onoda, and K.-R. M{\"u}ller, ``Soft margins for adaboost,''
  \emph{Machine learning}, vol.~42, no.~3, pp. 287--320, 2001.

\bibitem{ren2015faster}
S.~Ren, K.~He, R.~Girshick, and J.~Sun, ``Faster r-cnn: Towards real-time
  object detection with region proposal networks,'' in \emph{Advances in neural
  information processing systems}, 2015, pp. 91--99.

\bibitem{he2017mask}
K.~He, G.~Gkioxari, P.~Doll{\'a}r, and R.~Girshick, ``Mask r-cnn,'' in
  \emph{Proceedings of the IEEE international conference on computer vision},
  2017, pp. 2961--2969.

\bibitem{redmon2016you}
J.~Redmon, S.~Divvala, R.~Girshick, and A.~Farhadi, ``You only look once:
  Unified, real-time object detection,'' in \emph{Proceedings of the IEEE
  conference on computer vision and pattern recognition}, 2016, pp. 779--788.

\bibitem{redmon2017yolo9000}
J.~Redmon and A.~Farhadi, ``Yolo9000: better, faster, stronger,'' in
  \emph{Proceedings of the IEEE conference on computer vision and pattern
  recognition}, 2017, pp. 7263--7271.

\bibitem{redmon2018yolov3}
------, ``Yolov3: An incremental improvement,'' \emph{arXiv preprint
  arXiv:1804.02767}, 2018.

\bibitem{bochkovskiy2020yolov4}
A.~Bochkovskiy, C.-Y. Wang, and H.-Y.~M. Liao, ``Yolov4: Optimal speed and
  accuracy of object detection,'' \emph{arXiv preprint arXiv:2004.10934}, 2020.

\bibitem{zhan2019esir}
F.~Zhan and S.~Lu, ``Esir: End-to-end scene text recognition via iterative
  image rectification,'' in \emph{Proceedings of the IEEE Conference on
  Computer Vision and Pattern Recognition}, 2019, pp. 2059--2068.

\bibitem{xie2019aggregation}
Z.~Xie, Y.~Huang, Y.~Zhu, L.~Jin, Y.~Liu, and L.~Xie, ``Aggregation
  cross-entropy for sequence recognition,'' in \emph{Proceedings of the IEEE
  Conference on Computer Vision and Pattern Recognition}, 2019, pp. 6538--6547.

\bibitem{wan2020vocabulary}
Z.~Wan, J.~Zhang, L.~Zhang, J.~Luo, and C.~Yao, ``On vocabulary reliance in
  scene text recognition,'' in \emph{Proceedings of the IEEE/CVF Conference on
  Computer Vision and Pattern Recognition}, 2020, pp. 11\,425--11\,434.

\bibitem{qiao2020seed}
Z.~Qiao, Y.~Zhou, D.~Yang, Y.~Zhou, and W.~Wang, ``Seed: Semantics enhanced
  encoder-decoder framework for scene text recognition,'' in \emph{Proceedings
  of the IEEE/CVF Conference on Computer Vision and Pattern Recognition}, 2020,
  pp. 13\,528--13\,537.

\bibitem{shannon1948mathematical}
C.~E. Shannon, ``A mathematical theory of communication,'' \emph{The Bell
  system technical journal}, vol.~27, no.~3, pp. 379--423, 1948.

\bibitem{luck1994spatial}
S.~J. Luck and S.~A. Hillyard, ``Spatial filtering during visual search:
  evidence from human electrophysiology.'' \emph{Journal of Experimental
  Psychology: Human Perception and Performance}, vol.~20, no.~5, p. 1000, 1994.

\bibitem{badcock1990low}
J.~C. Badcock, F.~A. Whitworth, D.~R. Badcock, and W.~J. Lovegrove,
  ``Low-frequency filtering and the processing of local—global stimuli,''
  \emph{Perception}, vol.~19, no.~5, pp. 617--629, 1990.

\bibitem{maragos1987morphological}
P.~Maragos and R.~Schafer, ``Morphological filters--part ii: Their relations to
  median, order-statistic, and stack filters,'' \emph{IEEE Transactions on
  acoustics, speech, and signal processing}, vol.~35, no.~8, pp. 1170--1184,
  1987.

\bibitem{otsu1979threshold}
N.~Otsu, ``A threshold selection method from gray-level histograms,''
  \emph{IEEE transactions on systems, man, and cybernetics}, vol.~9, no.~1, pp.
  62--66, 1979.

\bibitem{rais2004adaptive}
N.~B. Rais, M.~S. Hanif, and I.~A. Taj, ``Adaptive thresholding technique for
  document image analysis,'' in \emph{8th International Multitopic Conference,
  2004. Proceedings of INMIC 2004.}\hskip 1em plus 0.5em minus 0.4em\relax
  IEEE, 2004, pp. 61--66.

\bibitem{hastie2009multi}
T.~Hastie, S.~Rosset, J.~Zhu, and H.~Zou, ``Multi-class adaboost,''
  \emph{Statistics and its Interface}, vol.~2, no.~3, pp. 349--360, 2009.

\bibitem{DBLP:journals/corr/TianHHH016}
\BIBentryALTinterwordspacing
Z.~Tian, W.~Huang, T.~He, P.~He, and Y.~Qiao, ``Detecting text in natural image
  with connectionist text proposal network,'' \emph{CoRR}, vol. abs/1609.03605,
  2016. [Online]. Available: \url{http://arxiv.org/abs/1609.03605}
\BIBentrySTDinterwordspacing

\bibitem{shi2017detecting}
B.~Shi, X.~Bai, and S.~Belongie, ``Detecting oriented text in natural images by
  linking segments,'' in \emph{Proceedings of the IEEE Conference on Computer
  Vision and Pattern Recognition}, 2017, pp. 2550--2558.

\bibitem{deng2018pixellink}
D.~Deng, H.~Liu, X.~Li, and D.~Cai, ``Pixellink: Detecting scene text via
  instance segmentation,'' \emph{arXiv preprint arXiv:1801.01315}, 2018.

\bibitem{zhou2017east}
X.~Zhou, C.~Yao, H.~Wen, Y.~Wang, S.~Zhou, W.~He, and J.~Liang, ``East: an
  efficient and accurate scene text detector,'' in \emph{Proceedings of the
  IEEE conference on Computer Vision and Pattern Recognition}, 2017, pp.
  5551--5560.

\bibitem{liao2016textboxes}
M.~Liao, B.~Shi, X.~Bai, X.~Wang, and W.~Liu, ``Textboxes: A fast text detector
  with a single deep neural network,'' \emph{arXiv preprint arXiv:1611.06779},
  2016.

\bibitem{kipf2016semi}
T.~N. Kipf and M.~Welling, ``Semi-supervised classification with graph
  convolutional networks,'' \emph{arXiv preprint arXiv:1609.02907}, 2016.

\bibitem{shi2016end}
B.~Shi, X.~Bai, and C.~Yao, ``An end-to-end trainable neural network for
  image-based sequence recognition and its application to scene text
  recognition,'' \emph{IEEE transactions on pattern analysis and machine
  intelligence}, vol.~39, no.~11, pp. 2298--2304, 2016.

\bibitem{shi2016robust}
B.~Shi, X.~Wang, P.~Lyu, C.~Yao, and X.~Bai, ``Robust scene text recognition
  with automatic rectification,'' in \emph{Proceedings of the IEEE conference
  on computer vision and pattern recognition}, 2016, pp. 4168--4176.

\bibitem{liu2018fots}
X.~Liu, D.~Liang, S.~Yan, D.~Chen, Y.~Qiao, and J.~Yan, ``Fots: Fast oriented
  text spotting with a unified network,'' in \emph{Proceedings of the IEEE
  conference on computer vision and pattern recognition}, 2018, pp. 5676--5685.

\bibitem{liao2018rotation}
M.~Liao, Z.~Zhu, B.~Shi, G.-s. Xia, and X.~Bai, ``Rotation-sensitive regression
  for oriented scene text detection,'' in \emph{Proceedings of the IEEE
  conference on computer vision and pattern recognition}, 2018, pp. 5909--5918.

\bibitem{bartz2017stn}
C.~Bartz, H.~Yang, and C.~Meinel, ``Stn-ocr: A single neural network for text
  detection and text recognition,'' \emph{arXiv preprint arXiv:1707.08831},
  2017.

\bibitem{liu2016ssd}
W.~Liu, D.~Anguelov, D.~Erhan, C.~Szegedy, S.~Reed, C.-Y. Fu, and A.~C. Berg,
  ``Ssd: Single shot multibox detector,'' in \emph{European conference on
  computer vision}.\hskip 1em plus 0.5em minus 0.4em\relax Springer, 2016, pp.
  21--37.

\bibitem{zhang2004boosting}
G.~Zhang, X.~Huang, S.~Z. Li, Y.~Wang, and X.~Wu, ``Boosting local binary
  pattern (lbp)-based face recognition,'' in \emph{Chinese Conference on
  Biometric Recognition}.\hskip 1em plus 0.5em minus 0.4em\relax Springer,
  2004, pp. 179--186.

\bibitem{king2009dlib}
D.~E. King, ``Dlib-ml: A machine learning toolkit,'' \emph{The Journal of
  Machine Learning Research}, vol.~10, pp. 1755--1758, 2009.

\bibitem{zhang2020resnest}
H.~Zhang, C.~Wu, Z.~Zhang, Y.~Zhu, Z.~Zhang, H.~Lin, Y.~Sun, T.~He, J.~Mueller,
  R.~Manmatha \emph{et~al.}, ``Resnest: Split-attention networks,'' \emph{arXiv
  preprint arXiv:2004.08955}, 2020.

\bibitem{tan2019efficientnet}
M.~Tan and Q.~V. Le, ``Efficientnet: Rethinking model scaling for convolutional
  neural networks,'' \emph{arXiv preprint arXiv:1905.11946}, 2019.

\bibitem{wang2018learning}
Y.~Wang, V.~I. Morariu, and L.~S. Davis, ``Learning a discriminative filter
  bank within a cnn for fine-grained recognition,'' in \emph{Proceedings of the
  IEEE conference on computer vision and pattern recognition}, 2018, pp.
  4148--4157.

\bibitem{krizhevsky2012imagenet}
A.~Krizhevsky, I.~Sutskever, and G.~E. Hinton, ``Imagenet classification with
  deep convolutional neural networks,'' in \emph{Advances in neural information
  processing systems}, 2012, pp. 1097--1105.

\bibitem{simonyan2014very}
K.~Simonyan and A.~Zisserman, ``Very deep convolutional networks for
  large-scale image recognition,'' \emph{arXiv preprint arXiv:1409.1556}, 2014.

\bibitem{szegedy2015going}
C.~Szegedy, W.~Liu, Y.~Jia, P.~Sermanet, S.~Reed, D.~Anguelov, D.~Erhan,
  V.~Vanhoucke, and A.~Rabinovich, ``Going deeper with convolutions,'' in
  \emph{Proceedings of the IEEE conference on computer vision and pattern
  recognition}, 2015, pp. 1--9.

\bibitem{he2016deep}
K.~He, X.~Zhang, S.~Ren, and J.~Sun, ``Deep residual learning for image
  recognition,'' in \emph{Proceedings of the IEEE conference on computer vision
  and pattern recognition}, 2016, pp. 770--778.

\bibitem{huang2017densely}
G.~Huang, Z.~Liu, L.~Van Der~Maaten, and K.~Q. Weinberger, ``Densely connected
  convolutional networks,'' in \emph{Proceedings of the IEEE conference on
  computer vision and pattern recognition}, 2017, pp. 4700--4708.

\bibitem{zhang2014part}
N.~Zhang, J.~Donahue, R.~Girshick, and T.~Darrell, ``Part-based r-cnns for
  fine-grained category detection,'' in \emph{European conference on computer
  vision}.\hskip 1em plus 0.5em minus 0.4em\relax Springer, 2014, pp. 834--849.

\bibitem{branson2014bird}
S.~Branson, G.~Van~Horn, S.~Belongie, and P.~Perona, ``Bird species
  categorization using pose normalized deep convolutional nets,'' \emph{arXiv
  preprint arXiv:1406.2952}, 2014.

\bibitem{lin2015bilinear}
T.-Y. Lin, A.~RoyChowdhury, and S.~Maji, ``Bilinear cnn models for fine-grained
  visual recognition,'' in \emph{Proceedings of the IEEE international
  conference on computer vision}, 2015, pp. 1449--1457.

\bibitem{article}
H.~Zhang, B.~Dong, B.~Feng, F.~Yang, and B.~Xu, ``Classification of financial
  tickets using weakly supervised fine-grained networks,'' \emph{IEEE Access},
  vol.~PP, pp. 1--1, 07 2020.

\bibitem{bell2016inside}
S.~Bell, C.~Lawrence~Zitnick, K.~Bala, and R.~Girshick, ``Inside-outside net:
  Detecting objects in context with skip pooling and recurrent neural
  networks,'' in \emph{Proceedings of the IEEE conference on computer vision
  and pattern recognition}, 2016, pp. 2874--2883.

\bibitem{hu2018squeeze}
J.~Hu, L.~Shen, and G.~Sun, ``Squeeze-and-excitation networks,'' in
  \emph{Proceedings of the IEEE conference on computer vision and pattern
  recognition}, 2018, pp. 7132--7141.

\bibitem{zhu2016scene}
Y.~Zhu, C.~Yao, and X.~Bai, ``Scene text detection and recognition: Recent
  advances and future trends,'' \emph{Frontiers of Computer Science}, vol.~10,
  no.~1, pp. 19--36, 2016.

\bibitem{tian2016detecting}
Z.~Tian, W.~Huang, T.~He, P.~He, and Y.~Qiao, ``Detecting text in natural image
  with connectionist text proposal network,'' in \emph{European conference on
  computer vision}.\hskip 1em plus 0.5em minus 0.4em\relax Springer, 2016, pp.
  56--72.

\bibitem{wu1992fast}
S.~Wu and U.~Manber, ``Fast text searching: allowing errors,''
  \emph{Communications of the ACM}, vol.~35, no.~10, pp. 83--91, 1992.

\bibitem{kim2014convolutional}
Y.~Kim, ``Convolutional neural networks for sentence classification,''
  \emph{arXiv preprint arXiv:1408.5882}, 2014.

\bibitem{liu2016recurrent}
P.~Liu, X.~Qiu, and X.~Huang, ``Recurrent neural network for text
  classification with multi-task learning,'' \emph{arXiv preprint
  arXiv:1605.05101}, 2016.

\bibitem{lai2015recurrent}
S.~Lai, L.~Xu, K.~Liu, and J.~Zhao, ``Recurrent convolutional neural networks
  for text classification,'' in \emph{Twenty-ninth AAAI conference on
  artificial intelligence}, 2015.

\bibitem{yang2016hierarchical}
Z.~Yang, D.~Yang, C.~Dyer, X.~He, A.~Smola, and E.~Hovy, ``Hierarchical
  attention networks for document classification,'' in \emph{Proceedings of the
  2016 conference of the North American chapter of the association for
  computational linguistics: human language technologies}, 2016, pp.
  1480--1489.

\bibitem{devlin2018bert}
J.~Devlin, M.-W. Chang, K.~Lee, and K.~Toutanova, ``Bert: Pre-training of deep
  bidirectional transformers for language understanding,'' \emph{arXiv preprint
  arXiv:1810.04805}, 2018.

\end{thebibliography}
\bibliographystyle{IEEEtran}
\vspace{12pt}
\end{document}